\documentclass[acmtog]{acmart}

\usepackage{booktabs} 

\setcitestyle{square}

\usepackage[ruled]{algorithm2e} 

\SetAlFnt{\small}
\SetAlCapFnt{\small}
\SetAlCapNameFnt{\small}
\SetAlCapHSkip{0pt}

\usepackage[skip=4pt]{caption}
\setcopyright{acmcopyright}

\begin{document}

\acmJournal{TOG}
\acmYear{2019}
\acmVolume{38}
\acmNumber{6}
\acmArticle{185}
\acmMonth{11}
\acmDOI{10.1145/3355089.3356574}

\title{Artistic Glyph Image Synthesis via One-Stage Few-Shot Learning}

\author{Yue Gao}
\authornote{Denotes equal contribution}
\email{gerry@pku.edu.cn}
\affiliation{%
	\institution{Wangxuan Institute of Computer Technology, Peking University}
	\city{Beijing}
	\country{China}
}

\author{Yuan Guo}
\authornotemark[1]
\email{guo-yuan@pku.edu.cn}
\affiliation{%
	\institution{Wangxuan Institute of Computer Technology, Peking University}
	\city{Beijing}
	\country{China}
}

\author{Zhouhui Lian}
\authornote{Corresponding author}
\email{lianzhouhui@pku.edu.cn}
\affiliation{%
	\institution{Wangxuan Institute of Computer Technology, Peking University}
	\city{Beijing}
	\country{China}
}

\author{Yingmin Tang}
\email{tangyingmin@pku.edu.cn}
\affiliation{%
	\institution{Wangxuan Institute of Computer Technology, Peking University}
	\city{Beijing}
	\country{China}
}

\author{Jianguo Xiao}
\email{xiaojianguo@pku.edu.cn}
\affiliation{%
	\institution{Wangxuan Institute of Computer Technology, Peking University}
	\city{Beijing}
	\country{China}
}

\begin{abstract}
Automatic generation of artistic glyph images is a challenging task that attracts many research interests. Previous methods either are specifically designed for shape synthesis or focus on texture transfer. In this paper, we propose a novel model, AGIS-Net, to transfer both shape and texture styles in one-stage with only a few stylized samples. To achieve this goal, we first disentangle the representations for content and style by using two encoders, ensuring the multi-content and multi-style generation. Then we utilize two collaboratively working decoders to generate the glyph shape image and its texture image simultaneously. In addition, we introduce a local texture refinement loss to further improve the quality of the synthesized textures. In this manner, our one-stage model is much more efficient and effective than other multi-stage stacked methods. We also propose a large-scale dataset with Chinese glyph images in various shape and texture styles, rendered from 35 professional-designed artistic fonts with 7,326 characters and 2,460 synthetic artistic fonts with 639 characters, to validate the effectiveness and extendability of our method. Extensive experiments on both English and Chinese artistic glyph image datasets demonstrate the superiority of our model in generating high-quality stylized glyph images against other state-of-the-art methods.
\end{abstract}

%
%


%
%


\begin{teaserfigure}
\label{fig_1}
  \centering
  \includegraphics[width=\linewidth]{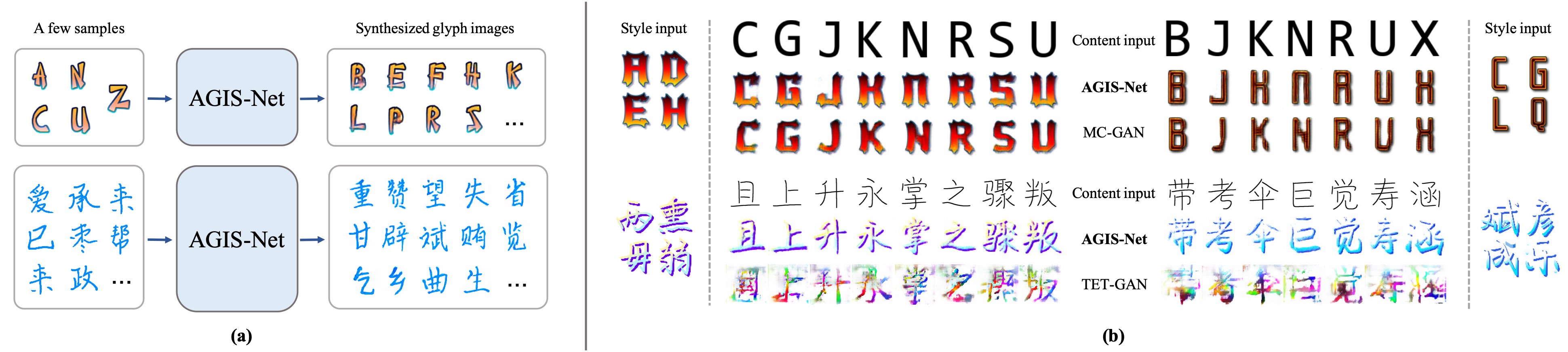}
  \caption{\label{fig:intro}
    (a) An overview of our method, given a few reference samples (5 for English or 30 for Chinese), glyph images of all other characters in the same style can be synthesized. (b) Examples of synthesized English/Chinese glyph images obtained by our proposed AGIS-Net, MC-GAN~\cite{azadi2018multi} and TET-GAN~\cite{yang2019tet}, respectively, please zoom in for better inspection.}
\end{teaserfigure}

\maketitle

\section{Introduction}

Artistic font design is a time-consuming and labor-intensive process due to the complex combination of lines, serif details, colors and textures. Moreover, maintaining a coherent style among all characters of a font is also difficult. In some language systems (e.g., Chinese, whose official character set GB18030 contains 27,533 characters), it is almost impossible to manually design all characters. Automatically generating glyph images for characters could be helpful for the design of artistic fonts. This paper focuses on the task of automatically synthesizing novel glyph images for characters in an artistic font using just a few samples created by a font designer (see Figure~\ref{fig:intro}(a)). Our method (AGIS-Net) allows to synthesize glyph images that are more coherent and realistic in terms of style and structure compared to state-of-the-art approaches (e.g., MC-GAN~\cite{azadi2018multi} and TET-GAN~\cite{yang2019tet}, see Figure \ref{fig:intro}(b)).

Up to now, a number of works have been reported for glyph image/glyph generation. They can be roughly classified into two groups, glyph shape synthesis and texture transfer. The first group mainly concentrates on the generation of geometric outlines: Campbell and Kautz \shortcite{campbell2014learning} built a font manifold, based on finding accurate correspondences between Latin glyphs' outlines, which typically fails when the number and/or complexity of glyphs greatly increase. Lian et al.~\shortcite{lian2018easyfont} attempted to extract strokes from given Chinese glyphs and learn to write corresponding strokes for other characters in the same style. But it is unsuited for many other scenarios such as Latin glyph synthesizing. For the second group, optimization-based methods~\cite{Yang2017AwesomeTS, Men2018ACF} have shown promising results. Based on the shape outline, patches extracted from the given glyph image with texture effects are rearranged to appropriate positions on the target glyph image. However, they require the outline shape image as reference to establish accurate correspondences and time-consuming online iterative optimization.

Recently, deep neural networks have spurred lots of interests in glyph image synthesis. With the development of Variational Autoencoders (VAEs)~\cite{kingma2013auto} and Generative Adversarial Networks (GANs)~\cite{goodfellow2014generative}, many methods have been proposed either for glyph shape image generation~\cite{zhang2018separating, Jiang2017DCFontAE, jiang2019scfont, guo2018creating} or for texture effects transfer~\cite{yang2019tet}. MC-GAN~\cite{azadi2018multi} is the first work that reports an end-to-end solution which is capable of learning styles of both shape outline and texture effects. It uses a stacked architecture that consists of two models. The first model tries to generate glyph shapes, and then the generated glyph shape image is fed into the second model to decorate textures on it. The results are impressive but there still exist some fatal drawbacks as follows: 1) The input content is limited to 26 Latin capital letters, and it is impossible to generalize the architecture to handle other writing systems (e.g., Chinese) that contain large numbers of characters. 2) The amount of parameters is extremely large as it has to input and output all 26 glyph images in one forward pass. Since the architectures of these two models are similar, there should exist some ways to reduce computational costs and storage requirements.

To effectively solve the above-mentioned problems, this paper proposes a novel model, \emph{\textbf{AGIS-Net}} \footnote{Source code and dataset are available at \url{https://hologerry.github.io/AGIS-Net/}} (Artistic Glyph Image Synthesis Network), which is capable of generating stylized glyph images by training on a small number of reference samples. To the best of our knowledge, our work is the first to transfer both shape and texture styles to arbitrarily large numbers of characters and generate high-quality synthesis results. Compared to the work~\cite{azadi2018multi} most relevant to ours, the proposed model has a wider application scenario, which is not only suitable for English but also Chinese and any other writing systems. Furthermore, unlike MC-GAN~\cite{azadi2018multi}, the proposed AGIS-Net is a one-stage model, which means that the generator directly outputs stylized glyph images with user-specified contents and styles.

More specifically, we use two encoders to extract content and style features separately, and apply two parallel decoders to recover glyph shape style and texture style. Namely, there exist two branches in our generator. The first branch acts as a guidance to the second branch and the two branches are implemented simultaneously. Since the two branches share the same content and style features from the encoders, less parameters in the model are needed. Furthermore, we use the contextual loss and introduce the local texture refinement loss to further improve image quality. The contextual loss~\cite{mechrez2018contextual}, similar to the perceptual loss, computes the gap between the ground truth and outputs at the feature level and is capable of improving the realism of the whole image. The local texture refinement loss aims at synthesizing high-quality local patches. It attempts to refine the details of the image by adversarial training. To verify the effectiveness and extendibility of our method, we also build a new Chinese artistic glyph image dataset which consists of 1,571,940 glyph images, containing 2,460 synthetic artistic font styles. Additionally, we collect 256,410 glyph images rendered from 35 professional-designed fonts, and each font has 7,326 characters. Qualitative and quantitative experiments conducted on both English and Chinese glyph image datasets demonstrate that our proposed method markedly outperforms existing approaches, synthesizing realistic and high-quality stylized glyph images.

In summary, our key contributions are listed as follows:
\begin{itemize}
	\item We propose a simple yet effective model, AGIS-Net, exploiting two parallel encoder-decoder branches, to transfer artistic font style with respect to both shape style and texture style within a single stage.
	\item We introduce a novel and computationally efficient loss function called the local texture refinement loss, which is helpful to improve the quality of synthesis results in few-shot style transfer tasks.
	\item We construct a new Chinese glyph image dataset, which consists of more than $1.8$ million images covering $2,460$ synthetic artistic font styles and 35 artist-designed font styles.
	\item Extensive experiments clearly verify the effectiveness and extendibility of our proposed method on few-shot learning, and demonstrate its superiority to other existing methods on artistic font style transfer.
\end{itemize}

\section{Related Work}

\subsection{Font Generation}
Campbell and Kautz~\shortcite{campbell2014learning} built a font manifold and generated new fonts by interpolation in a high dimensional space. Lian et al.~\shortcite{lian2018easyfont} proposed a system to automatically generate large-scale Chinese handwriting fonts by learning styles of stroke shape and layout separately. Balashova et al.~\shortcite{Balashova2019LearningAS} developed a stroke-based geometric model for glyph synthesis, embedding fonts on a manifold using purely geometric features. \cite{Baluja2016LearningTS} is one of the earliest works to use deep neural networks to generate English glyph images. Upchurch et al.~\shortcite{Upchurch2016FromAT} considered glyph image synthesis as an image analogy task and proposed a modified VAE to separate image style from content. Lyu et al.~\shortcite{Lyu2017AutoEncoderGG} proposed to apply an image-to-image translation model to learn mappings from glyph images in the standard font style to those with desired styles. Jiang et al.~\shortcite{jiang2019scfont} designed a deep stacked neural network with the guidance of glyph structure information to synthesize high-quality Chinese fonts.

\subsection{Style Transfer}
Style transfer aims at migrating a given image's style to another image while preserving the latter one's content. Gatys et al.~\shortcite{Gatys2015ANA} pioneered a style transfer scheme based on Convolutional Neural Networks (CNNs), getting quite appealing results. Johnson et al.~\shortcite{Johnson2016PerceptualLF} extended the work~\cite{Gatys2015ANA} by introducing the perceptual loss for training and using feed-forward networks for image transformation. More recently, Huang and Belongie~\shortcite{Huang2017ArbitraryST} presented an effective approach that for the first time enables arbitrary style transfer in real-time. Li et al.~\shortcite{Li2017UniversalST} embedded a pair of patch-based feature transforms, whitening and coloring, to an image reconstruction network to synthesize styled images with high visual quality. Gu et al. \shortcite{Gu2018ArbitraryST} took advantage of reshuffling deep features to achieve arbitrary style transfer while preserving local style patterns and preventing artifacts.

There also exist some works focusing on texture transfer of glyph images. For example, Yang et al.~\shortcite{Yang2017AwesomeTS} explored the task of generating special text effects for typography by proposing an optimization-based model. Based on the traditional texture transfer technique, Men et al.~\shortcite{Men2018ACF} proposed to adopt structure information to effectively guide the synthesis process. More recently, a deep learning based method was reported~\cite{yang2019tet} to accelerate the transfer process while maintaining the image quality.

In contrast to the above-mentioned works, many researchers intend to explicitly disentangle the content and style of images. For instance, Tenenbaum et al.~\shortcite{Tenenbaum2000SeparatingSA} presented a general bilinear model to solve two-factor tasks, providing sufficiently expressive representations of factor interactions. Wang et al.~\shortcite{Wang2016GenerativeIM} proposed to factorize the image generation process and use two GANs for surface normal map and image generation, respectively. Gonzalez-Garcia et al.~\shortcite{GonzalezGarcia2018ImagetoimageTF} introduced the concept of cross-domain disentanglement and separated the internal representation into a shared part and two exclusive parts. Kazemi et al.~\shortcite{Kazemi2018StyleAC} proposed a method to train GANs to learn disentangled style and content representations of the data. Motivated by these works, we also disentangle the content and style of glyph images to make our model capable of handling the tasks of precisely transferring both shape and texture styles, by using a new network architecture and several novel loss functions.

\subsection{Generative Adversarial Networks}
Generative Adversarial Networks (GANs) were originally proposed in~\cite{goodfellow2014generative} by introducing the adversarial process to generative models. Since then, many works~\cite{radford2015unsupervised,salimans2016improved} have been proposed to improve the performance of GANs. For instance, Conditional GANs~\cite{chen2016infogan,odena2017conditional} attempt to use labels or images to control the generation and have been applied in many application scenarios such as image-to-image transformation.

In~\cite{isola2017image}, Isola et al. proposed a unified image-to-image framework, \emph{Pix2Pix}, based on conditional GANs. Then, Zhu et al. \shortcite{zhu2017toward} proposed BicycleGAN that can model multimodal distribution and output diverse images. Mao et al.~\shortcite{MSGAN} presented a mode seeking regularization term to address the mode collapse problem for general GAN models. However, paired data are typically hard to obtain in many tasks. To solve the problem, several unpaired methods~\cite{zhu2017unpaired,lee2018diverse} have been reported whose key idea is to use the cycle consistency loss in training. Also, Iizuka et al.~\shortcite{Iizuka2017GloballyAL} proposed to use global and local discriminators to solve the image completion problem. More recently, there exist some works, such as~\cite{Cloutre2019FIGRFI} and~\cite{Liu2019FewShotUI}, which explore the few-shot learning based image generation task.

The work most relevant to our model is MC-GAN~\cite{azadi2018multi} which regards glyph image synthesis as an image-to-image translation task, and aims at transforming a content image to the stylized glyph image. MC-GAN connects two networks in series to get a two-stage model to achieve the goals of shape style transfer and texture style transfer, respectively.


\section{Method Description}\label{sec:method}

\begin{figure*}[t!]
	\label{fig_2}
	\centering
	\includegraphics[width=0.9\linewidth]{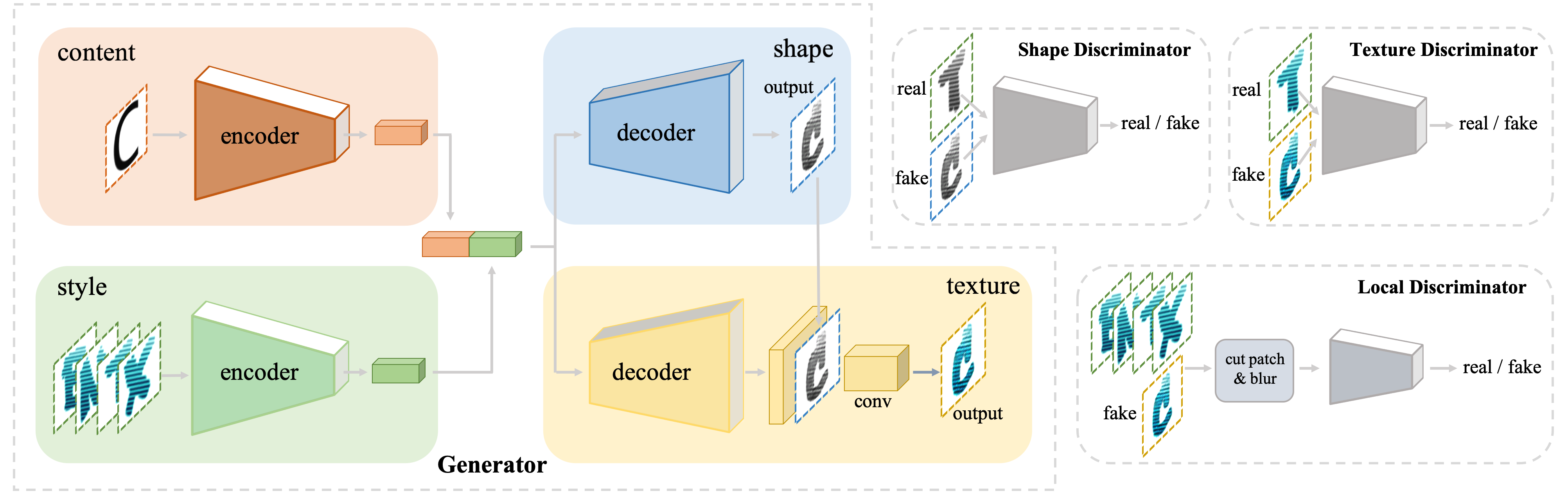}
	\caption{\label{fig:network}
		The architecture of our model, \emph{AGIS-Net}, which is composed of one Generator and three Discriminators. Line arrows denote the data stream, and the Generator's shape and texture output images are fed into the Shape Discriminator and Texture Discriminator, respectively. The real samples of the Shape Discriminator are the gray-scale version of real samples of the Texture Discriminator. In the mean time, the style input images and the texture output image are fed into the Local Discriminator. }
\end{figure*}

As mentioned above, our goal is to synthesize stylized glyph images with conditional contents and styles. Similar to other GANs, we have a generator and several discriminators in our model. To obtain better performance for this task, we specifically design the network architecture and loss function, which will be explicitly discussed in Section~\ref{sec:net} and \ref{sec:loss}.

We formulate the generation process as a mapping from a content reference image $x_{c}$ and a small set of style reference images $\mathcal{X}_{s} = \left\{x_{1}, x_{2}, x_{3}, ...\right\}$, which all have the same style but different contents, to the output glyph image $y$ with the content as $x_{c}$ and style as $\mathcal{X}_{s}$. The content image $x_{c}$ is a binary glyph image in a standard font style (e.g., \textit{Code New Roman} for English or an average font style for Chinese) containing little style information. The reason for using a set of stylized images rather than just one as our style reference is that each stylized glyph image is composed of its content and style, and thus we have to find a way to disentangle the style from the image. Given a set of stylized glyph images instead of one, our model could be able to extract the common feature from them, namely style information, and ignore the content. Assume that there are $n$ stylized images in our few-shot reference set: $\mathcal{R}_{s} = \left\{r_{1}, r_{2}, ..., r_{n}\right\}$. During each forward propagation, we randomly select a set of $m$ ($m < n$) images from $\mathcal{R}_{s}$ and concatenate them together in a channel-wise manner as our style reference input: $\mathcal{X}_{s} \subset \mathcal{R}_{s}$. The reasons for doing this are twofold: 1) The model can have many different combinations of $m$ images as input, making it more robust. In contrast, if we simultaneously feed all these $n$ reference samples, the style input will always be the same during training. Thus, the model could easily fall into a local optimum, losing generalization ability. 2) For other writing systems (e.g., Chinese, $n = 30$), feeding all these $n$ samples at the same time will dramatically increase the model size.

For stable convergence, we pre-train our model before implementing few-shot learning. After that, we can fine-tune our pre-trained model to any specific artistic style as we want. In this paper, we use the Chinese glyph image dataset created by us and the English glyph image dataset proposed in~\cite{azadi2018multi} for pre-training.

\begin{figure}[t!]
	\label{fig_3}
	\centering
	\includegraphics[width=0.85\columnwidth]{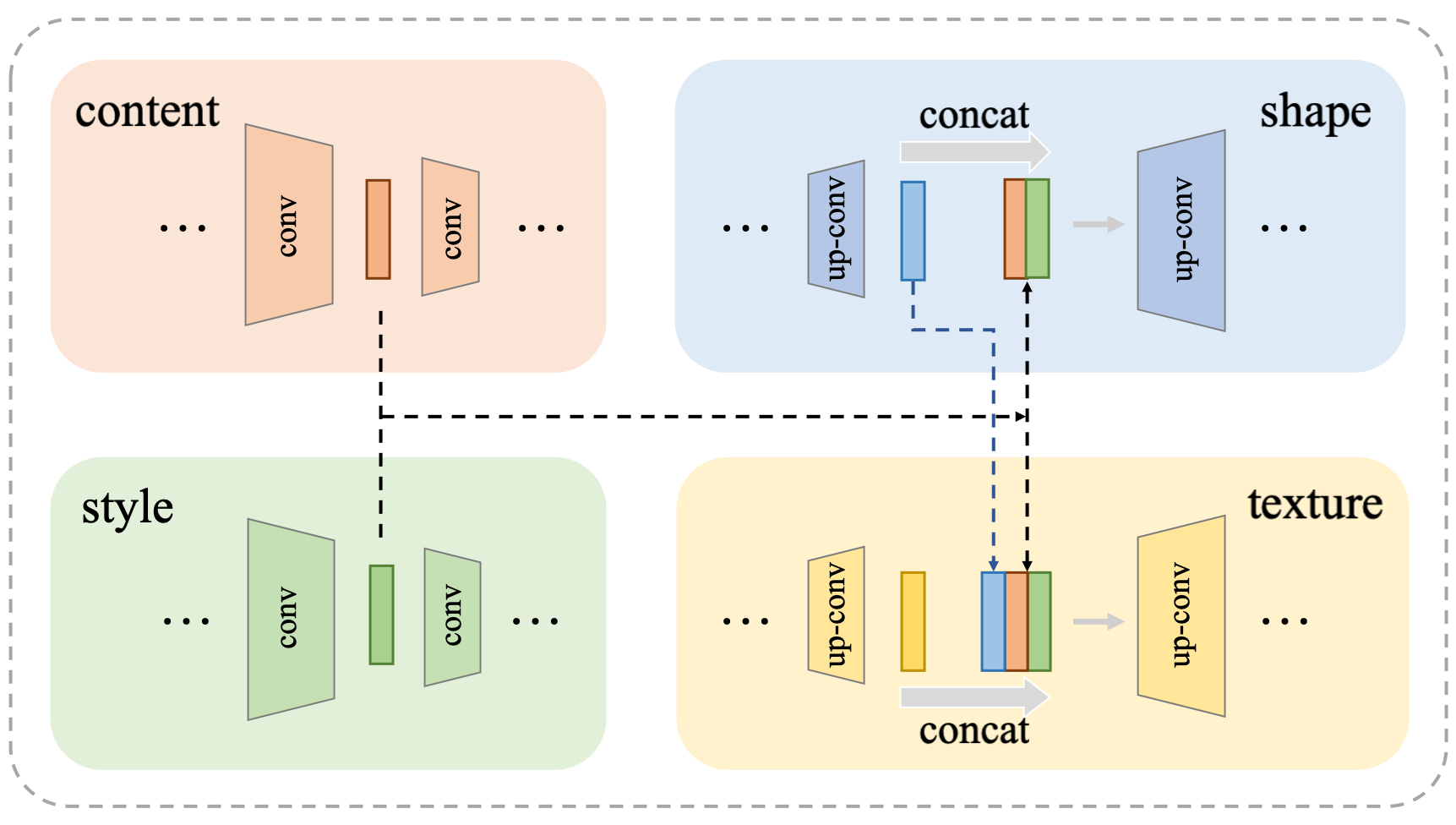}
	\caption{\label{fig:skipconnection}
		Illustration of skip connections in the middle layers of the generator. The last layer of texture decoder is omitted here. Dashed line arrows and line arrows denote data stream. Block arrows with texts mean operations. } 
\end{figure}

\subsection{Network Architecture}\label{sec:net}
As shown in Figure~\ref{fig:network}, our model consists of a Generator $G$ and three discriminators: Shape Discriminator $D_{sha}$, Texture Discriminator $D_{tex}$ and Local Discriminator $D_{local}$.

The two encoders separately extract the content and style features of the input. They have almost the same structure that consists of several convolution layers, except for the number of input channels. The two decoders behave differently: one for glyph shape and the other for texture. Inspired by Pix2Pix~\cite{isola2017image}, we propose to use skip connections in the two encoder-decoder branches, as shown in Figure~\ref{fig:skipconnection}, so that these two encoder-decoder branches can work together. For the shape decoder, there are several up-convolution layers. The input of each layer is a concatenation of the features of the previous layer and the corresponding layer in two encoders. Then, a gray-scale shape image $y_{gray}$ can be generated from the shape decoder. For the texture decoder, it is similar but the input of each layer is extra concatenated by the features of the corresponding shape decoder layer. At the end of the texture decoder, there is one more convolution layer fed with a concatenation of previous features and the gray-scale image $y_{gray}$, so that all information from the shape decoder can be shared with the texture decoder.

The purpose of using this special skip-connection architecture is that features at different scales are all important. For example, higher-level features contain more abstract style information while lower-level features contain more specific style information, we intend to exploit them as much as possible letting the model learn effective and sufficient information. We propose to use two separated decoders in our model mainly because, compared to color and texture, the shape is much more variable in glyph images. With such kind of network architecture, our model can pay more attention to the shape style.

For adversarial training, there are 3 discriminators in our model. The first two, $D_{sha}$ and $D_{tex}$, are PatchGAN~\cite{isola2017image} like discriminators. The two decoders' outputs $y_{gray}$ and $y$ are fed into these two discriminators as fake samples, respectively. The third one, $D_{local}$, is used for refining local textures. Details of all discriminators will be described in Section \ref{sec:loss}.

\begin{figure}[t!]
	\centering
	\includegraphics[width=0.85\columnwidth]{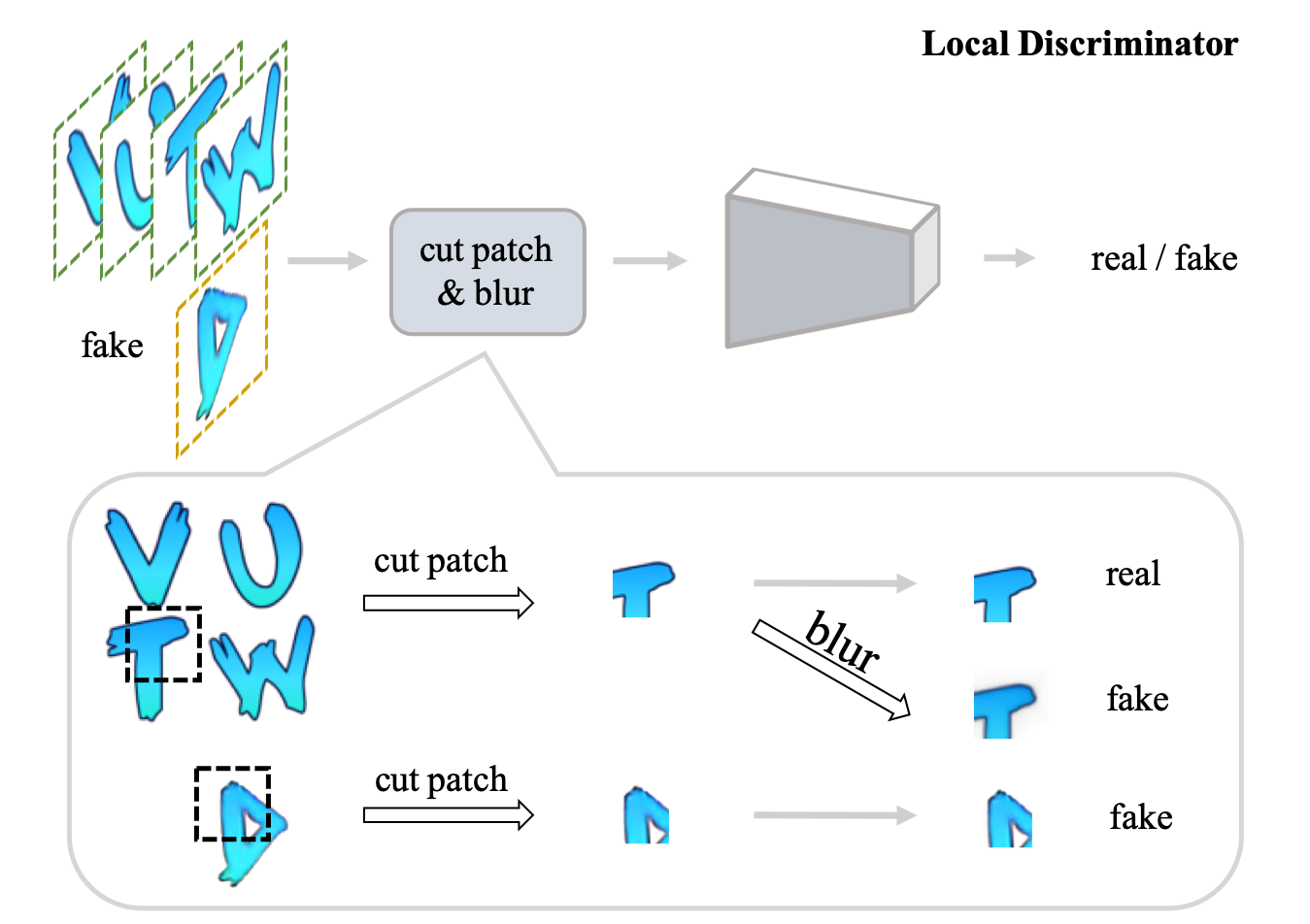}
	\caption{\label{fig:localdis}
		Illustration of the Local Discriminator. The block arrows denote operations and line arrows are data stream without any operation. In the cut patch \& blur module, we randomly cut several patches as our positive and negative samples, and also apply the Gaussian Blur on positive samples to get fuzzy negative samples.}
\end{figure}

\subsection{Objective Function}
\label{sec:loss}
The objective function of our model consists of four terms: adversarial loss, $L_1$ loss, contextual loss and local texture refinement loss
\begin{equation}
L(G) = L_{adv} + L_{1} + L_{CX} + L_{local}.
\end{equation}
\label{adv_train}
\noindent\textbf{Adversarial loss:} Similar to most GANs, we impose a standard adversarial game to train the generator $G$ and discriminators $D_{sha}$, $D_{tex}$. As mentioned above, $D_{sha}$ is trained for gray-scale image $y_{gray}$ and $D_{tex}$ for texture image $y$
\begin{gather}
	\begin{align}
		L(D_{sha}) = &\mathbb{E}_{t^{r}_{gray}}  [\log(D_{sha}   (t^{r}_{gray}) )] + \nonumber\\
		&\mathbb{E}_{y_{gray}}  [\log(1-D_{sha} (y_{gray})     )],
	\end{align}
\end{gather}
\begin{gather}
	\begin{align}
		L(D_{tex}) = &\mathbb{E}_{t^{r}_{s}}  [\log(D_{tex}   (t^{r}_{s})    )] + \nonumber\\
		&\mathbb{E}_{y}  [\log(1-D_{tex} (y)  )],
	\end{align}
\end{gather}
\begin{gather}
	\begin{align}
		L_{adv} = & \lambda_{adv\_sha}\mathbb{E}_{y_{gray}}[\log(1-D_{sha} (y_{gray}) )] + \nonumber\\
		& \lambda_{adv\_tex}\mathbb{E}_{y}[\log(1-D_{tex} (y) )],
	\end{align}
\end{gather}
where $y$ and $y_{gray}$ are machine-generated images, $t^{r}_{s}$ is a real glyph image with texture effects, $t^{r}_{gray}$ denotes the gray-scale version of $t^{r}_{s}$, $\lambda_{adv\_sha}$ and $\lambda_{adv\_tex}$ are weights for balancing these terms. If we train the model for a character which is in the few-shot reference set $\mathcal{R}_{s}$, $t^{r}_{s}$ will be the ground truth image of $y$, if not it will be randomly selected from the style reference input set $\mathcal{X}_{s}$.

\noindent\textbf{$L_1$ loss:} To stabilize our training, we use an $L_1$ loss in our objective function. The $L_1$ loss function also has two terms, respectively for gray-scale images and texture images, which are defined as
\begin{gather}
	\begin{align}
		L_{1gray} = & \mathbb{E}_{y_{gray}, \hat{y}_{gray}} \parallel y_{gray} - \hat{y}_{gray} \parallel_1, \\
		L_{1tex}  = & \mathbb{E}_{y, \hat{y}} \parallel y -        \hat{y} \parallel_1, \\
        	L_{1} = &\lambda_{L_{1gray}} L_{1gray} + \lambda_{L_{1tex}} L_{1tex},
	\end{align}
\end{gather}
where $\lambda_{L_{1gray}}$ and $\lambda_{L_{1tex}}$ are weights, $\hat{y}$ and $\hat{y}_{gray}$ are ground truth images. It is worth explaining that during training we explore all glyphs in the font library, which means we synthesize glyph images for all characters. For example, if we train on the English dataset, we generate glyph images for all 26 capital letters, although most of them might not belong to the few-shot reference set. When implementing few-shot learning (i.e., fine-tuning on a specific stylized font), if the glyph is in the few-shot reference set, we have ground truth images and the weights will not be 0. Otherwise, if ground truth images are unavailable, these two weights should be set to 0. Although unseen characters (i.e., out of the few-shot reference set) cannot contribute to the $L_1$ loss, they are still useful for the adversarial training, helping us get more satisfactory results. The real samples for discriminators $D_{sha}$ and $D_{tex}$ are chosen in two different manners depending on whether the characters belong to $\mathcal{R}_{s}$ or not. In the stage of pre-training, as all glyphs have corresponding ground truth images, the weights of $L_1$ will never be zero.

\noindent\textbf{Contextual loss:} The contextual loss was recently proposed in \cite{mechrez2018contextual}. It is a new and effective way to measure the similarity between two images, requiring no spatial alignment. As the spatial alignment is required for the $L_1$ loss, if the synthesized image is not exactly spatially aligned to the ground truth image (e.g., a small displacement or rotation), the $L_1$ loss will be high but the synthesis result is often visually acceptable. The contextual loss leads the model to pay more attention to style features at a high level, not just differences in pixel values. Therefore, we regard the contextual loss as a complementary to the $L_1$ loss.

Here we briefly explain the contextual loss. The key idea of contextual loss is to treat an image as a collection of features, and measure the similarity between two images based on the similarity of the feature map collections, while ignoring the spatial alignment of features. Given an image $x$ and its target image $y$, we gather the feature maps (e.g., VGG19~\cite{Simonyan2015VeryDC} features): $X=\left\{x_1,x_2, ..., x_N\right\}$ and $Y=\left\{y_1, y_2,..., y_N\right\}$. For each $y_j$ in $Y$, we find the most similar $x_i$ to it, calculate the distance between them, and convert it to similarity. Then we calculate the average similarity over all similarity values and apply the negative logarithm to get the loss value. Formally, it is defined as

\begin{equation}
    L_{CX}(x, y) = - log (CX(X, Y)),
\end{equation}
\begin{equation}
    CX(X, Y) = \frac{1}{N}\sum_{j}\max_{i}CX_{ij},
\end{equation}
where $CX_{ij}$ is the similarity between $x_i$ and $y_j$. To get $CX_{ij}$, firstly, we calculate the cosine distance $d_{ij}$ between $x_i$ and $y_j$, and then normalize the distances, shift from distances to similarities by exponentiation and normalize the similarities, which are defined by the following equations
\begin{gather}
	\begin{align}
	\hat{d}_{ij} = & \frac{d_{ij}}{\min_{k}d_{ik}+\epsilon},\\
	w_{ij} = & \exp\frac{1-\hat{d}_{ij}}{h},\\
	CX_{ij} = & w_{ij} / \sum_{k}w_{ik},
	\end{align}
\end{gather}
where $\epsilon$ and $h$ are hyper-parameters, and we fix them to $\epsilon = 1e-5$ and $h=0.5$ as the original paper~\cite{mechrez2018contextual}.

In our model, we apply the contextual loss to both the gray-scale image $y_{gray}$ and texture image $y$ by computing
\begin{gather}
	\begin{align}
		L_{CX_{gray}} = & - \frac{1}{L}  \sum_{l} log(CX(\Phi^{l}(y_{gray}), \Phi^{l}(\hat{y}_{gray}))), \\
		L_{CX_{tex}}  = & - \frac{1}{L}  \sum_{l} log(CX(\Phi^{l}(y),        \Phi^{l}(\hat{y}))),\\
	       L_{CX} = &\lambda_{CXgray}L_{CXgray} + \lambda_{CXtex}L_{CXtex},
	\end{align} 
\end{gather}
where $\Phi^{l}(\cdot)$ means extracted features from the $l$th layer of VGG19, $L$ is the number of used layers, $\lambda_{CXgray}$ and $\lambda_{CXtex}$ are weights. Since we apply the contextual loss to the output and its ground truth, $\lambda_{CXgray}$ and $\lambda_{CXtex}$ will be zero for the adversarial training of unseen characters, same as $\lambda_{L_{1gray}}$ and $\lambda_{L_{1tex}}$ in the $L_1$ loss.

\noindent\textbf{Local texture refinement loss:} With such a small number of glyph images used in few-shot learning, the positive and negative samples are highly unbalanced. Therefore, we propose the local texture refinement loss to address this problem.

As shown in Figure~\ref{fig:localdis}, we randomly cut patches from images, and feed them to the Local Discriminator instead of using the whole image. In this manner, training samples will be relatively sufficient and balanced. Moreover, to get better texture details, we manually blur some positive samples with a Gaussian Filter and regard them as negative samples when training. Through this blurring operation, we can build a bridge between real and fuzzy samples, so that $\textit{D}_{local}$ will force the Generator to synthesize more realistic images with less artifacts and noise. Obviously, the above-mentioned operations only involve a small amount of calculations, and $\textit{D}_{local}$ is also computationally efficient since the size of patches it processes is small. Although the idea of using image patches is similar to PatchGAN~\cite{isola2017image}, our motivation and implementing details are quite different: 1) We use patches to alleviate the problem of insufficient training samples, while PatchGAN aims to penalize structure at the scale of patches with a score indicating real/fake of a patch; 2) The way to build a bridge between real and fuzzy samples is novel and effective for this task.

The patches are used for training the third discriminator $\textit{D}_{local}$, whose negative samples consist of generated patches and blurred patches. So the loss function is defined as
\begin{gather}
	\begin{align}
		L(D_{local}) = & \mathbb{E}_{p_{real}}[\log(D_{local}   (p_{real}) )] + \nonumber\\
		& \mathbb{E}_{p_{blur}}[\log(1-D_{local} (p_{blur}) )] + \nonumber \\
		& \mathbb{E}_{p_y}[\log(1-D_{local}      (p_{y})    )], \\
	L_{local} = & \lambda_{local} \mathbb{E}_{p_y}[\log(1-D_{local}(p_{y}))],
	\end{align}
\end{gather}
where $p_{real}$ and $p_{y}$ represent the patches from style reference input images and generated images, respectively, $p_{blur}$ means the blurred patches of $p_{real}$, and $\lambda_{local}$ denotes the balancing weight.

Finally, the proposed model can be trained by playing the following minimax game
\begin{equation}
\min_{G}\max_{D_{sha},D_{tex},D_{local}} L(G, D_{sha}, D_{tex}, D_{local}).
\end{equation}


\section{Experiments}


\begin{figure}[t]
	\label{fig_5}
	\centering
	\includegraphics[width=0.95\columnwidth]{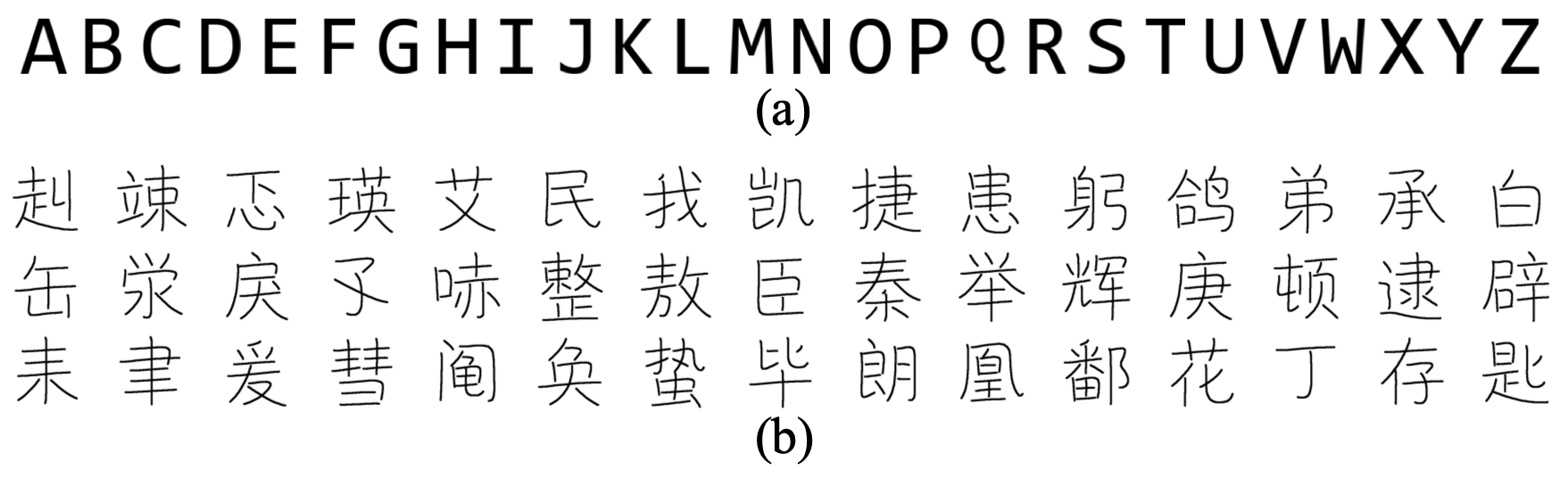}
	\caption{\label{fig:content}
		Examples of content input glyph images on the English and Chinese datasets.}
\end{figure}

\subsection{Experimental Settings}
\subsubsection{Datasets}
As mentioned in Section \ref{sec:method}, for a specific character, the input content reference image is identical for different artistic styles. As shown in Figure \ref{fig:content}(a), we use the content input in the same font style (i.e., \textit{Code New Roman}) as MC-GAN~\cite{azadi2018multi} for the English dataset; for Chinese, as shown in Figure \ref{fig:content}(b), we adopt the commonly used average font style~\cite{jiang2019scfont, guo2018creating}. The principle of choosing the font style of content input is that the shape style it contains should be as common as possible.

\begin{figure}[t]
	\label{fig_6}
	\centering
	\includegraphics[width=0.95\columnwidth]{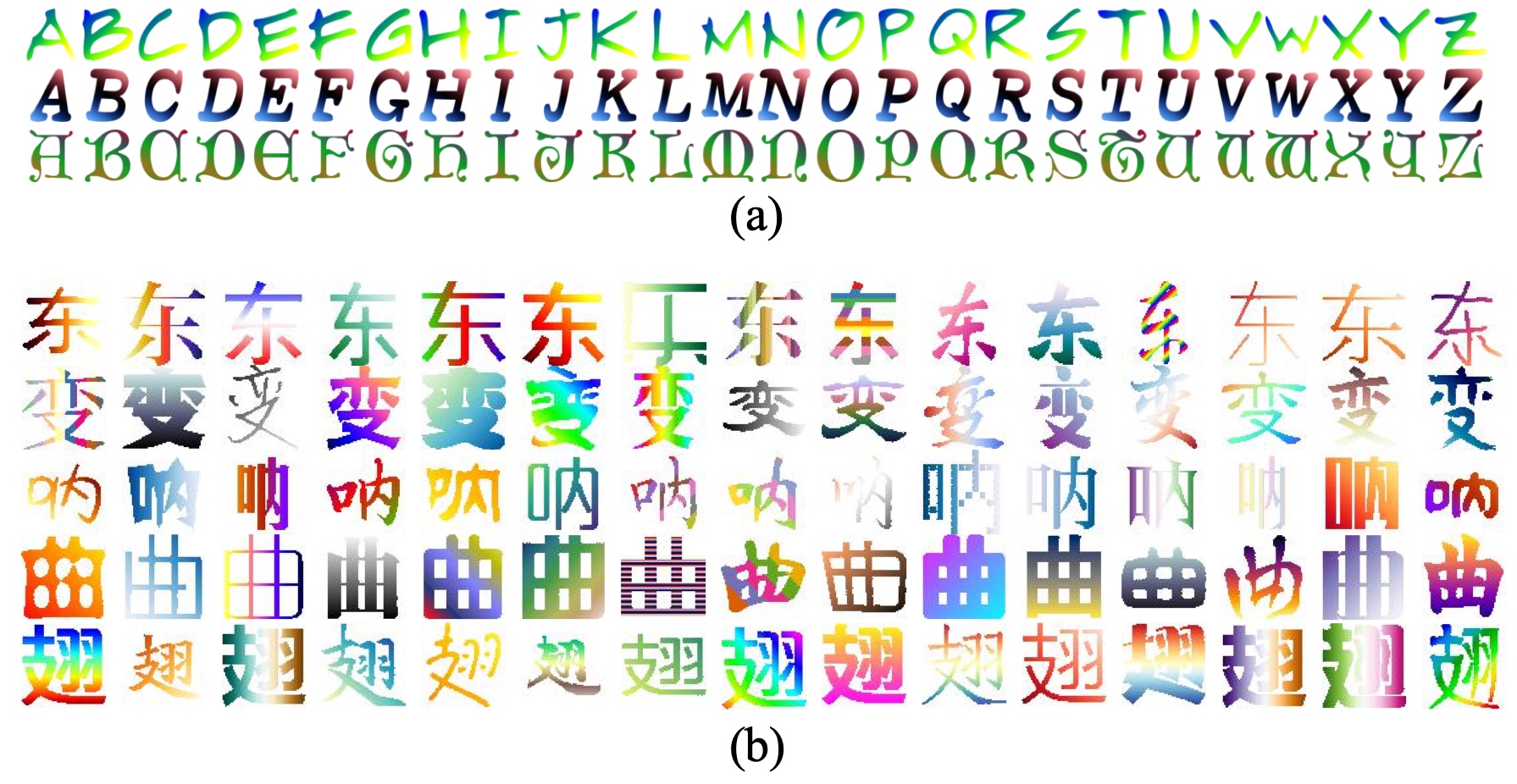}
	\caption{\label{fig:data}
		Examples of synthetic artistic glyph images. (a) The English glyph image dataset~\cite{azadi2018multi}, (b) Our Chinese glyph image dataset.}
\end{figure}

We use the English glyph image dataset, proposed by~\cite{azadi2018multi}, which contains 32,046 synthetic artistic fonts, each with 26 glyphs as shown in Figure \ref{fig:data}(a), to pre-train our model. The same test set as MC-GAN is used to fine-tune our model in few-shot learning, which contains 35 professional-designed English fonts with special text effects. Note that the English glyph image dataset only contains 26 capital letters.

Moreover, we also build a new publicly-available Chinese glyph image dataset for our experiments to verify our model's extendibility. For the pre-training set, as shown in Figure~\ref{fig:data}(b), we first render glyph images for 639 representative Chinese characters in 246 normal Chinese font styles. Then, to convert them into glyph images with textures, we apply gradient colors and various stripe textures on the original binary images. Specifically, 10 different kinds of gradient colors or stripe textures are applied to each font style. Overall, the dataset contains 1,571,940 different artistic glyph images. For few-shot learning, we select 35 artist-designed fonts with textures as the test set and each font consists of 7,326 Chinese characters.

\subsubsection{Implementation details}
In our experiments, we have six convolution layers~\cite{Krizhevsky2012ImageNetCW} and six up-convolution (transposed convolution) layers~\cite{Dumoulin2016AGT} in the generator of the proposed AGIS-Net, each layer is equipped with Instance Normalization~\cite{ulyanov2016instance} and ReLU~\cite{Nair2010RectifiedLU}. We follow the structure of discriminators in Pix2Pix~\cite{isola2017image} to design our three discriminators, which output score maps instead of a single value. All images are with the size $64\times64\times3$ except for the patch fed into $\textit{D}_{local}$ that is $32\times32\times3$. Weights in the loss function are selected as $\lambda_{adv\_sha}=1.0$, $\lambda_{adv\_tex}=1.0$, $\lambda_{L_{1gray}}=50.0$, $\lambda_{L_{1tex}}=100.0$, $\lambda_{CXgray}=15.0$, $\lambda_{CXtex}=25.0$ and $\lambda_{local}=1.0$, keeping unchanged during two training stages.

In pre-training, we train 20 epochs for the English glyph image dataset and 10 for the Chinese dataset. The batch size is 100 for both. When implementing few-shot learning (i.e., fine-tuning on a specific artistic font style), for English glyph images in each font style, we use the batch size 26 with 3,000 training epochs and validate the model every 50 epochs. We explore the 26 capital letters in the English dataset, most of these characters do not have ground truth images during few-shot learning. For Chinese glyph images in each font style, we use the batch size 100 with 500 training epochs, exploring 500 characters in the dataset, similarly, only a few of them have ground truth images. 

\subsubsection{Competitors}
We compare our method with four recently proposed image-to-image translation approaches. The first two are leading general-purpose image-to-image translation models, and the last two are state-of-the-art artistic font style transfer models. We directly use the source codes and default settings provided by the authors for these methods which are briefly described as follows.
\begin{itemize}
	\item BicycleGAN~\cite{zhu2017toward}: BicycleGAN learns a multi-modal mapping between two image domains, and can output diverse images from one input image.
	\item MS-Pix2Pix~\cite{MSGAN}: Mode Seeking GAN (MSGAN) utilizes a novel mode seeking regularization term to address the mode collapse issue for cGANs. We compare our method with the Pix2Pix based MSGAN, here we call it MS-Pix2Pix.
	\item MC-GAN~\cite{azadi2018multi}: MC-GAN uses a stacked conditional GAN to transfer both the shape and texture styles of glyph images, solving this challenging task for the first time and getting impressive results. However, MC-GAN can only handle 26 English capital letters and is hard to scale up.
	\item TET-GAN~\cite{yang2019tet}: TET-GAN consists of a stylization subnetwork and a destylization subnetwork. It learns to disentangle and recombine the content and style features of text effects images, through processes of style transfer and removal. TET-GAN uses the shape outline as a guidance to transfer the texture style only. For fair comparison, we directly apply their method on our datasets, meaning that the model is required to transfer the glyph shape style as well.
\end{itemize}

\subsubsection{Evaluation metrics}
For quantitative evaluation, we adopt four commonly-used metrics in many image generation tasks: Inception Score (IS)~\cite{salimans2016improved}, Fr\'echet Inception Distance (FID)~\cite{heusel2017gans}, structural similarity (SSIM) index and pixel-level accuracy (pix-acc). Specifically, IS is used to measure the realism and diversity of generated images. FID is employed to measure the distance between two distributions of synthesized glyph images and ground truth images, while SSIM aims to measure the structural similarity between them. Since FID and IS can not directly reflect the quality of synthesized character images, we also use the pixel-level accuracy (pix-acc) to evaluate performance. Higher values of IS, SSIM and pix-acc are better, whereas for FID, the lower the better.

\subsection{Pre-training Results}
\begin{figure}[t]
	\centering
	\includegraphics[width=0.95\columnwidth]{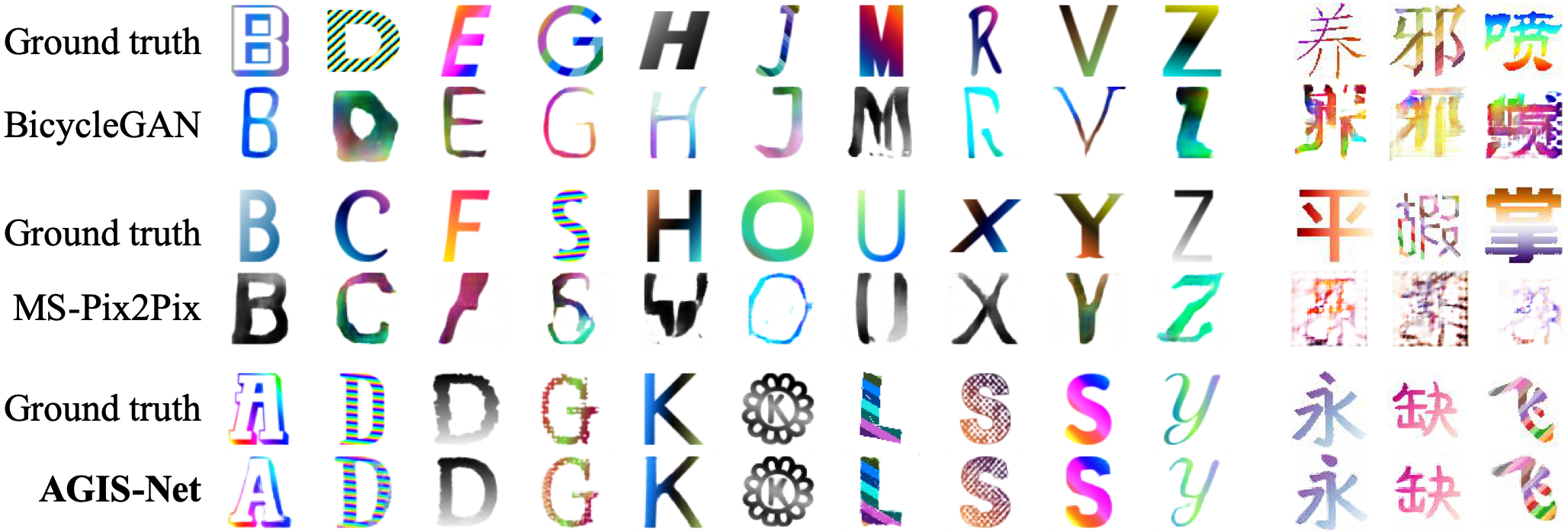}
	\caption{\label{fig:pre-train}
		Examples of some glyph images generated by BicycleGAN~\cite{zhu2017toward}, MS-Pix2Pix~\cite{MSGAN} and our AGIS-Net, on English and Chinese glyph image datasets during pre-training.}
\end{figure}

As mentioned above, we pre-train all models on the datasets with glyph images in synthetic artistic font styles. Figure~\ref{fig:pre-train} shows some results generated by our method, BicycleGAN~\cite{zhu2017toward} and MS-Pix2Pix~\cite{MSGAN}, respectively, during pre-training. Since we shuffle the input data after every epoch, the content and style may not be the same for all models in the same iteration. As we can see, although all explored characters' ground truth images are available during training, BicycleGAN and MS-Pix2Pix still can hardly learn the style information and often synthesize poor-quality glyph images. For BicycleGAN~\cite{zhu2017toward}, although it can capture the content information and output roughly correct glyph shapes for the English dataset, it can not process the style information well. The more recent work, MS-Pix2Pix~\cite{MSGAN}, can not even handle the content information well, and sometimes outputs glyph images with incorrect contents. What is worse, both of these two methods are prone to mode collapse, and can not even converge on the Chinese dataset which is more challenging. It is clear that our model's learning ability on this specific task is significantly better than the general-purpose image-to-image translation models (e.g., BicycleGAN and MS-Pix2Pix). Therefore, here we do not show results of these two models for the more challenging few-shot learning task, which can be found in the supplementary material.


\subsection{Parameter Studies} \label{sec-parameter}

Our goal is to generate novel glyph images for characters in an artistic font using just a few input samples. It is obvious that the few-shot size and style input size could affect the synthesizing performance. Therefore, we conduct extensive experiments to find an optimal setting. Here we examine the effects of different values of $n$, the size of few-shot reference set $\mathcal{R}_{s} = \left\{r_{1}, r_{2}, ..., r_{n}\right\}$, and $m$, the size of style input set $\mathcal{X}_{s} = \left\{x_{1}, x_{2}, ..., x_{m}\right\}$. All the few-shot reference sets for each artistic font style are randomly selected in this experiment.

\begin{figure*}[t]
	\centering
	\includegraphics[width=0.9\linewidth]{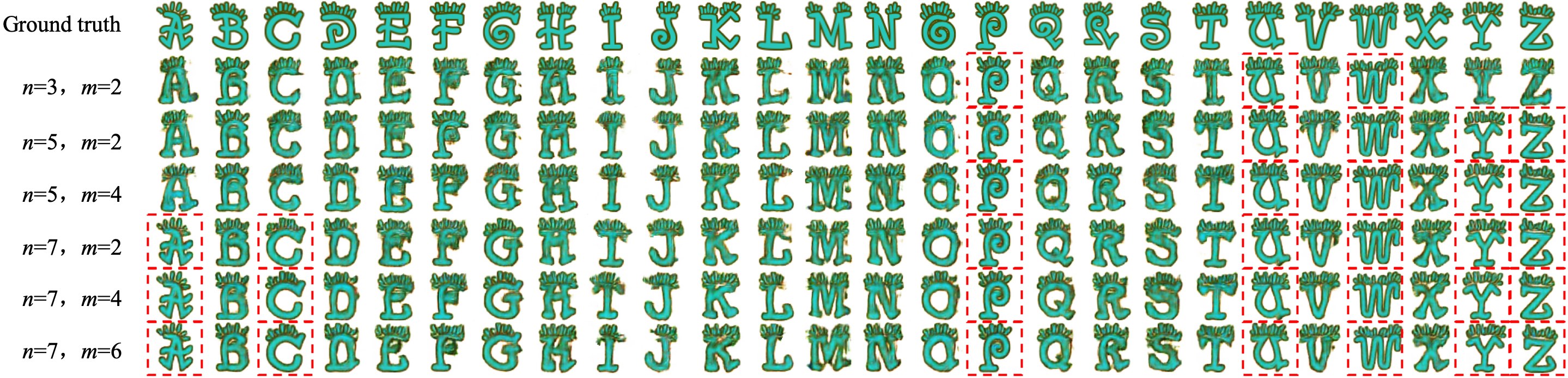}
	\caption{\label{fig:fewsize-e}
		Visual comparison of our models with different few-shot set size $n$ and style input size $m$ on the English glyph image dataset. Characters in the few-shot reference set are marked in red boxes. }
\end{figure*}

In Figure~\ref{fig:fewsize-e} and Table \ref{tab:fewsize-t}, we compare the performance of our methods with 6 pairs of different settings on the English dataset. From both the qualitative and quantitative results, we can see that with more training samples available, the quality of results synthesized by our method improves, with clearer contours and smaller fuzzy regions. Although just given very few samples, synthesis results are already visually pleasing. We can see that, for the style input size, the difference is not so obvious. But if the size is too small (e.g., $m=2$), results become less satisfactory. There is an interesting phenomenon that the Inception Scores of our methods on all cases are higher than the ground truth, which means the realism and diversity of ground truth are worst. This is obviously meaningless. So, the conceptions of realism and diversity are not suitable for this glyph image synthesizing task. Therefore, among all the four metrics, IS does not have much reference value for this task. It is also necessary to point out why the SSIM and pix-acc's relative values are small. For our task, we focus on generating glyph images in which there are lots of white pixels outside the glyph and thus a small change on those values may indicate dramatically large changes for the visual appearance (see Figure \ref{fig:fewsize-e} and Table \ref{tab:fewsize-t} where $n$ = 3, $m$ = 2 and $n$ = 5, $m$ = 4). Since parameters of the model are increasing with larger few-shot size and style input size, to balance the image quality and model size, we use the few-shot size $n=5$ and the style input size $m=4$ in the latter experiments for the English glyph image dataset.

\begin{figure}[t!]
	\centering
	\includegraphics[width=0.95\columnwidth]{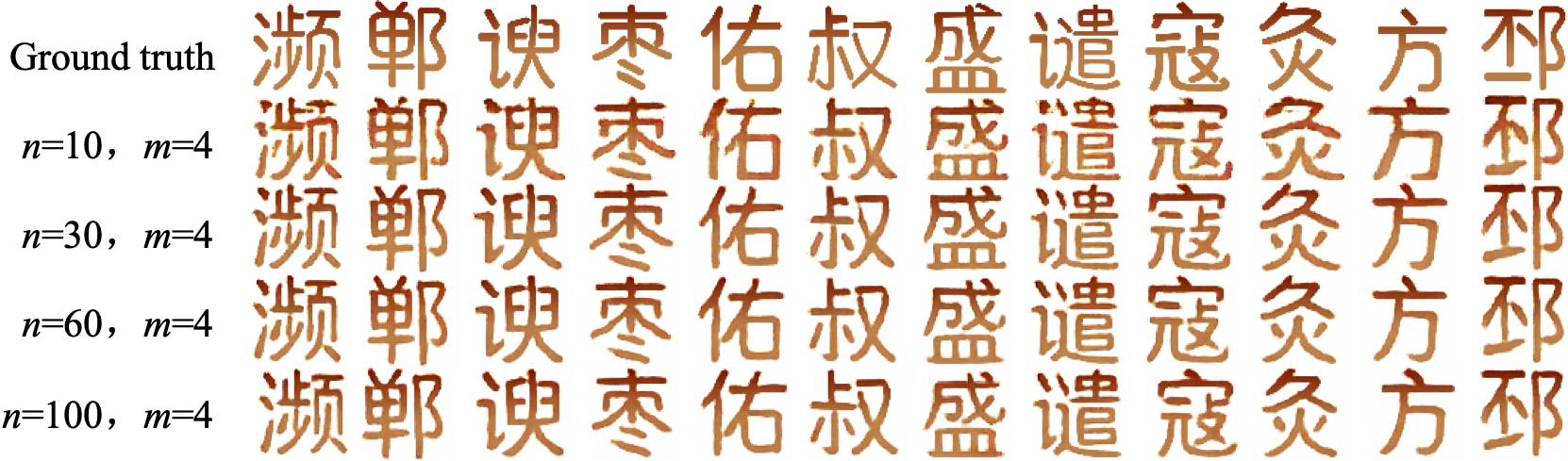}
	\caption{\label{fig:fewsize-c}
		Comparison of synthesis results obtained using our models with different values of few-shot size on the Chinese glyph image dataset. All these characters do not belong to the few-shot reference set.}
\end{figure}

Similar experiments are also conducted on the Chinese dataset. We compare the performance of our method with different few-shot sizes (i.e., $n=10$, 30, 60 and 100) and different style input sizes (i.e., $m=4$ and 8). Due to page limit, here we just show some synthesis results with the style input size $m=4$ in Figure \ref{fig:fewsize-c}, which demonstrates the effectiveness of the proposed method in synthesizing Chinese glyph images. Same as English, we also provide quantitative results on the Chinese dataset. As shown in Table \ref{tab:fewsize-t-cn}, we can see that the quality of synthesis results with $m=8$ are almost the same as that with $m=4$. Although the quality of results can be improved with more training samples, glyph images synthesized by our method with few-shot size 30 are already good enough for practical uses. Similarly, in the following experiments conducted on the Chinese dataset, the size of few-shot reference set and the style input size are fixed as $n=30$ and $m=4$, respectively.


\subsection{Effects of Input Data} \label{sec-input}

In this section, we analyze the influences of different content inputs and different style few-shot reference sets.

\subsubsection{Content input font style}

\begin{figure}[t!]
	\centering
	\includegraphics[width=0.95\columnwidth]{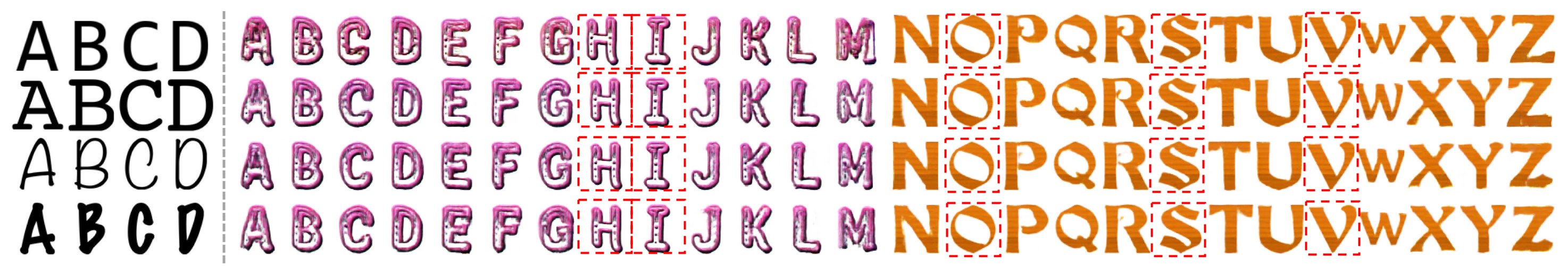}
	\caption{\label{fig:content_input}
		Our synthesis results using content inputs in different font styles, the left part shows some samples in those 4 different fonts.}
\end{figure}

As mentioned before, the content input is used to specify which character the model synthesizes. Along with the \textit{Code New Roman} used by MC-GAN, we select three extra fonts: \textit{Courier}, \textit{Noteworthy} and \textit{Marker Felt}. From the right part of Figure~\ref{fig:content_input}, we can see that the difference among glyph images of the same character synthesized using content inputs in different font styles are insignificant in general. This is mainly due to the fact that the content encoder is trained to extract content information while ignoring style information during both pre-training and fine-tuning procedures.

\subsubsection{Style few-shot reference set}

\begin{figure}[t!]
	\centering
	\includegraphics[width=0.95\columnwidth]{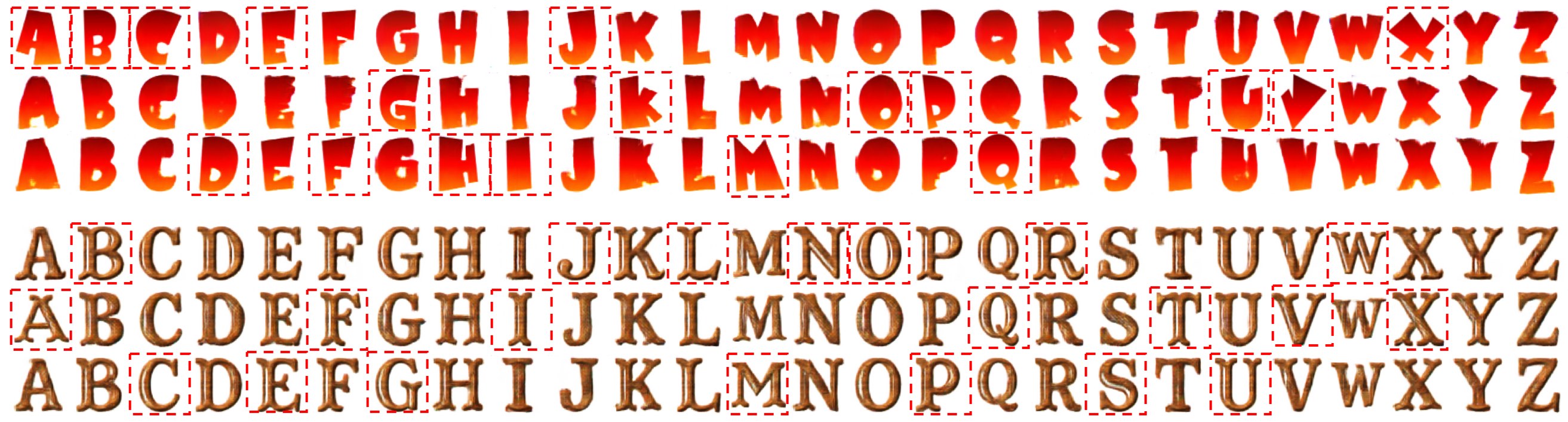}
	\caption{\label{fig:style_input}
		Our synthesis results using different style few-shot reference sets. For each artistic style, we select 3 different reference sets whose elements are marked in red boxes.}
\end{figure}

Apart from the same style few-shot reference set as MC-GAN, we randomly select two extra reference sets. As we can see, all results shown in Figure~\ref{fig:style_input} have consistent color and texture. For some characters in the first font, such as 'A', 'H', and 'Z', the shape styles of the results are inconsistent, which means different reference sets will lead to slightly different synthesized glyph shapes. However, for the second font, in which most characters share the same shape style, there is no marked difference between the results with different reference sets. It can be observed from this experiment that our style encoder can successfully extract the common feature for the given style few-shot reference set, especially for the font that all characters share the same shape style.

\begin{table}[t!]
	\centering
	\caption{Quantitative comparison of our models with different few-shot size $n$ and style input size $m$ on the English glyph image dataset.}
	\label{tab:fewsize-t}
	\begin{tabular}{lcccc}
		\toprule
		Setting           & IS              & FID             & SSIM            &  pix-acc \\
		\midrule
		$n$=3, $m$=2      & 3.3026          & 88.217          & 0.6648          &  0.6014  \\
		$n$=5, $m$=2      & 3.3111          & 80.277          & 0.7063          &  0.6139  \\
		$n$=5, $m$=4      & 3.2653          & 75.536          & 0.7219          &  0.6250  \\
		$n$=7, $m$=2      & 3.4046          & 74.567          & 0.7389          &  0.6241  \\
		$n$=7, $m$=4      & \textbf{3.4096} & 71.380          & \textbf{0.7470} &  0.6318  \\
		$n$=7, $m$=6      & 3.2221          & \textbf{69.979} & 0.7448          &  \textbf{0.6349} \\
		ground truth      & 3.0172          & -               & -               &  - \\
		\bottomrule
	\end{tabular}
\end{table}

\begin{table}[t!]
	\centering
	\caption{Quantitative comparison of our models with different few-shot size $n$ and style input size $m$ on the Chinese glyph image dataset.}
	\label{tab:fewsize-t-cn}
	\begin{tabular}{lcccc}
		\toprule
		Setting          & IS              & FID               & SSIM            & pix-acc  \\
		\midrule
		$n$=10, $m$=4    & 2.0015          & 90.545            & 0.5702          & 0.6825   \\
		$n$=10, $m$=8    & 1.9964          & 84.106            & 0.5711          & 0.6848   \\
		$n$=30, $m$=4    & \textbf{2.1122} & 70.875            & 0.6116          & 0.7035   \\
		$n$=30, $m$=8    & 2.0717          & 69.413            & 0.6118          & 0.7030   \\
		$n$=60, $m$=4    & 2.0467          & 64.072            & 0.6332          & 0.7137   \\
		$n$=60, $m$=8    & 2.0090          & 64.396            & 0.6347          & 0.7139   \\
		$n$=100, $m$=4   & 2.0019          & 58.839            & \textbf{0.6559} & \textbf{0.7216}   \\
		$n$=100, $m$=8   & 2.0173          & \textbf{58.623}   & 0.6558          & 0.7211   \\
		ground truth     & 1.8474          & -                 & -               & - \\
		\bottomrule
	\end{tabular}
\end{table}

\begin{figure*}[t!]
	\centering
	\includegraphics[width=0.85\linewidth]{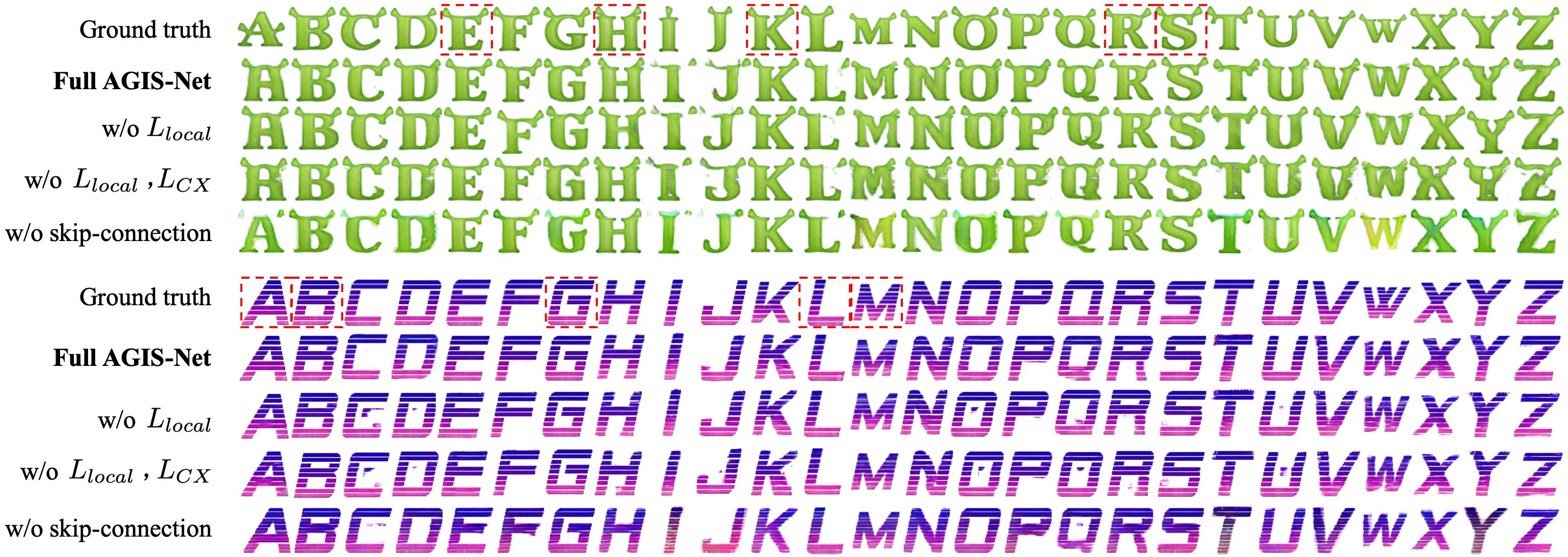}
	\caption{\label{fig:ablation}
		Ablation studies for the proposed AGIS-Net. Glyph images in the few-shot reference sets are marked in red boxes. The first to fifth rows show the ground truth, results of our full AGIS-Net, results of AGIS-Net without the local texture refinement loss, results of AGIS-Net without both the local texture fine loss and the contextual loss, and results of AGIS-Net without skip connections, respectively.}
\end{figure*}


\subsection{Ablation Studies} \label{sec-ablation}

In this section, we perform experiments to verify the effectiveness of each key component in our model. In Figure~\ref{fig:ablation}, we demonstrate the effects of the contextual loss, the local texture refinement loss and skip connections. Stylized glyph images in the fifth row are generated by the model without skip connections. Comparing the images in the second row synthesized by our full AGIS-Net with those in the fifth row, we can see that the glyph shape style can be handled well without skip connections, but not the texture style such as colors for both two fonts and white lines for the second font. Thereby, we conclude that the skip connections play an important role in capturing and recovering texture style information. Comparing synthesis results in the second and third rows, we can see how the local refinement loss performs. Taking 'J' and 'V' for the first font and 'C', 'D', 'O' and 'U' for the second font for example, synthesis texture details become poor and much noise appears without adopting the local texture refinement loss. The effectiveness of the contextual loss in this specific task can be verified by comparing synthesis results in the third and fourth rows. We can see that the contextual loss helps to maintain more shape outline style information, such as, 'A', 'D', 'N' and 'Z' in the first font and 'H', 'N' and 'S' in the second font.

\begin{table}
	\centering
	\caption{Quantitative results for ablation studies, w/o denotes without.}
	\label{tab:ablation}
	\begin{tabular}{lcccc}
		\toprule
		& IS              & FID             & SSIM            & pix-acc          \\
		\midrule
		w/o skip-connection          & 2.3365          & 109.555         & 0.6584          & 0.5876           \\
		w/o $L_{local}$ and $L_{CX}$ & 2.4002          & 92.457          & 0.6936          & 0.6158           \\
		w/o $L_{local}$              & 2.4283          & 80.994          & 0.7002          & 0.6169           \\
		\textbf{Full AGIS-Net}       & \textbf{3.2653} & \textbf{75.536} & \textbf{0.7219} & \textbf{0.6250}  \\
		\bottomrule
	\end{tabular}
\end{table}

\begin{figure}
	\centering
	\includegraphics[width=0.85\columnwidth]{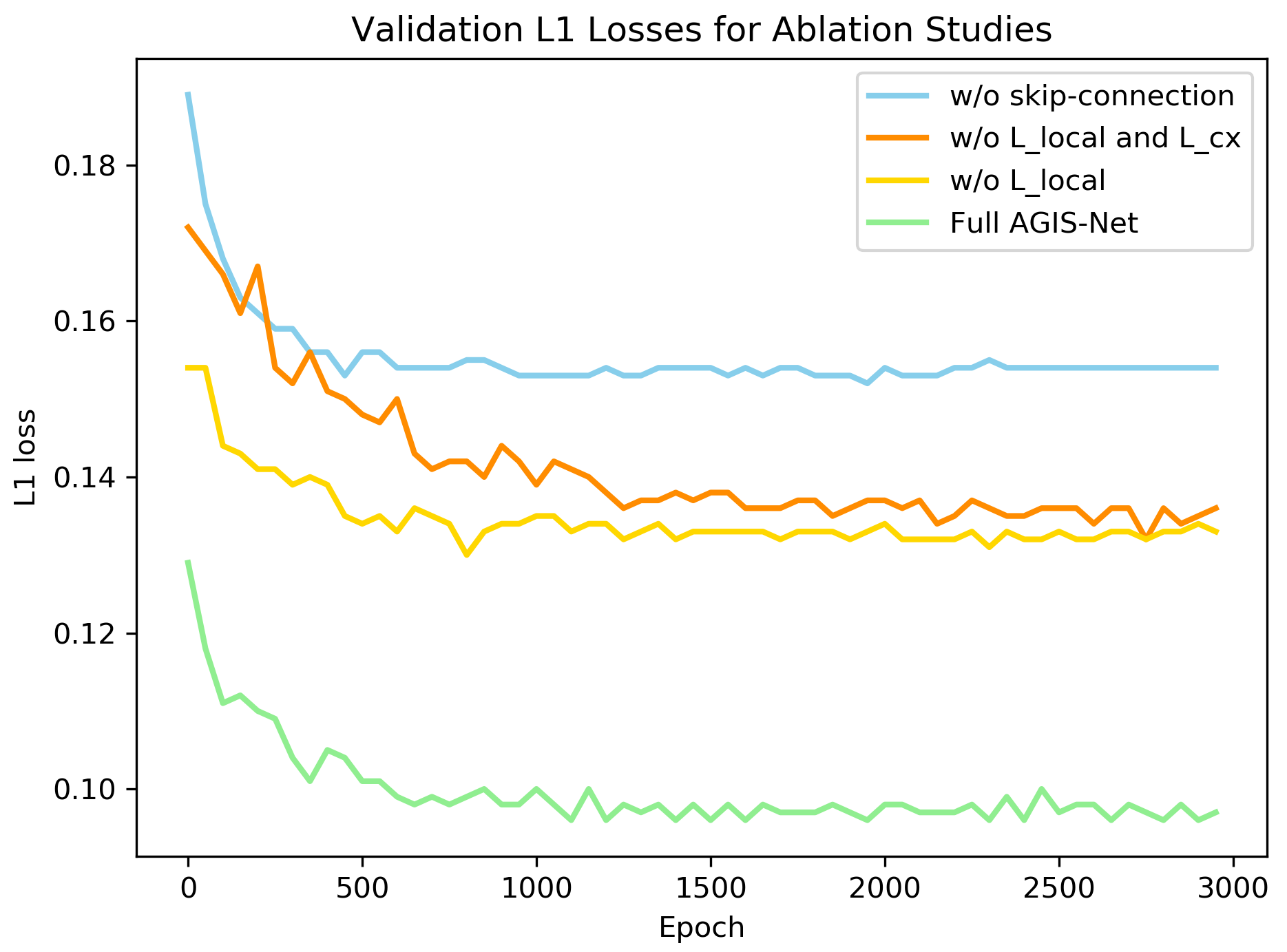}
	\caption{\label{fig:l1_loss}
		Validation $L_1$ losses for ablation studies on all glyph images during the fine-tuning stage.}
\end{figure}

Furthermore, we also provide quantitative results (see Table~\ref{tab:ablation}) evaluated on all 35 English artist-designed fonts, which intuitively reflect the influence of each component to the whole model. Values of these metrics clearly demonstrate that the novel skip connections in the Generator, which makes the two encoder-decoder branches work collaboratively, play the most important role in this task. Secondly, the local texture refinement loss also makes strong contribution in generating high-quality local texture details and noise-free synthesis results. Last but not the least, the contextual loss, which does not need spatial alignment, helps the model produce better shape outlines. From $L_1$ losses computed on the validation set during fine-tuning shown in Figure \ref{fig:l1_loss}, we can also get the same conclusion about the effects of those key components.

\begin{figure*}[t]
	\centering
	\includegraphics[width=0.85\linewidth]{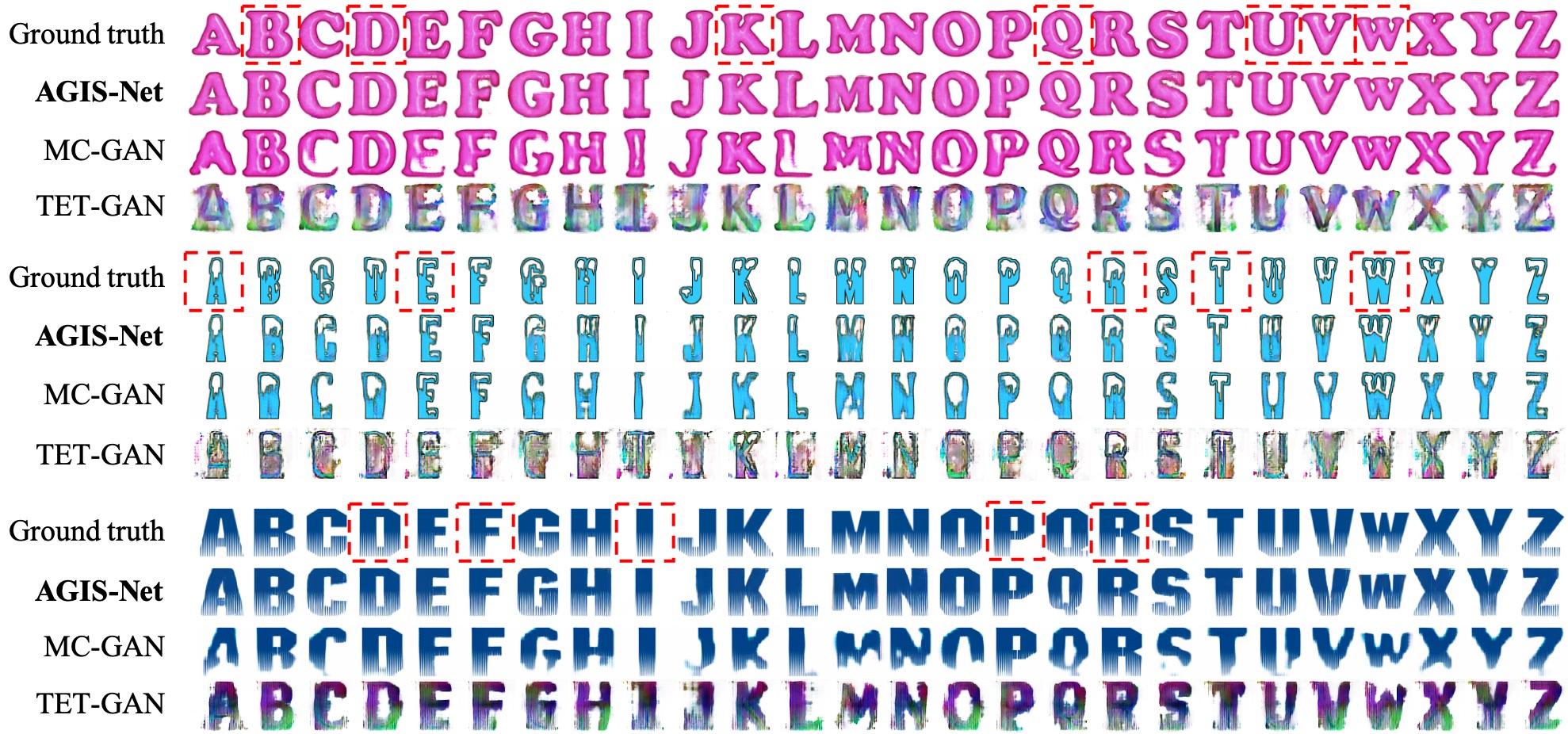}
	\caption{\label{fig:comparison}
		Visual comparison of glyph images synthesized by different methods for some English artist-designed fonts. Our AGIS-Net clearly outperforms MC-GAN~\cite{azadi2018multi} and TET-GAN~\cite{yang2019tet}, providing more complete shape outlines and uniform texture styles. }
\end{figure*}

\subsection{Comparison with the State of the Art}

In this section, we compare the performance of our model with other existing methods. Currently, the work that is most relevant to our method is MC-GAN~\cite{azadi2018multi} which also provides the state-of-the-art performance for stylized glyph image synthesis.
More recently, Yang et al. \shortcite{yang2019tet} proposed TET-GAN which shows promising results on the task of text effects transfer. But it requires the corresponding binary glyph image as input reference, unlike our task where the shape of the glyph is also synthesized. For fair comparison, we choose the same few-shot learning data as the default setting of the original MC-GAN~\cite{azadi2018multi} and fix the font style of input content images to the standard \textit{Code New Roman} in all three models. Then we generate the glyph images of all 26 characters in the above-mentioned 35 English artist-designed fonts using different methods and compare the performance of them both qualitatively and quantitatively.

\subsubsection{Visual comparison}
As shown in Figure \ref{fig:comparison}, our method clearly outperforms all other existing approaches, in both glyph shape and texture styles transfer. Moreover, our method could successfully decouple content and style, producing stylized glyph images based on the given contents and styles. As mentioned before, TET-GAN~\cite{yang2019tet} requires binary stylized glyph images as input reference, and thus it performs poorly under our experimental settings. MC-GAN~\cite{azadi2018multi} synthesizes reasonable results whose general shape and texture styles can be successfully transferred from input glyph images. But the details of images are not well synthesized, such as containing edge noise and incomplete shape outlines. 

\subsubsection{Quantitative evaluation}

\begin{table}
	\centering
	\caption{Quantitative comparison of our AGIS-Net, TET-GAN~\cite{yang2019tet} and MC-GAN~\cite{azadi2018multi} on the English glyph image dataset.}
	\label{tab:comparison}
	\begin{tabular}{lcccccc}
		\toprule
		Model               & IS               & FID             & SSIM            & pix-acc          & User prefer. \\
		\midrule
		TET-GAN             & 3.6378           & 193.043         & 0.4772          &  0.4322          &  0.0523      \\
		MC-GAN              & 3.7867           & 98.922          & 0.6620          &  0.5713          &  0.2463      \\
		\textbf{AGIS-Net}    & \textbf{3.8151}  & \textbf{73.893} & \textbf{0.7217} &  \textbf{0.6249} &  \textbf{0.7014}      \\
		\bottomrule
	\end{tabular}

\end{table}

Although the visual appearance is much more intuitive to reflect the quality of synthesis results in the image generation task, quantitative evaluation metrics can give a higher-level indication of performance on the whole dataset. Table~\ref{tab:comparison} shows the quantitative comparison of our method and other two approaches evaluated on the above-mentioned 35 English artist-designed fonts. In addition to the four evaluation measures, we also conduct a user study. Specifically, for each artistic font style we randomly select 5 characters which are not contained in the style few-shot reference set. Then for each character, a participant is asked to choose the one that possesses the best quality and has the most similar style as the reference set among the 3 glyph images synthesized by these 3 methods. 60 participants have finished all questions of this user study. Statistical results are shown in the 6th column of Table~\ref{tab:comparison}. We also list the number of parameters for all models in Table~\ref{tab:comparison-para}. We can see that our model performs much better while requiring fewer parameters than other state-of-the-art methods. We also observe that some general-purpose models, although having smaller amounts of parameters, such as BicycleGAN~\cite{zhu2017toward} and MS-Pix2Pix~\cite{MSGAN}, cannot perform well even on the pre-training datasets since they are basically unsuited for this specific task.

\begin{table}
	\centering
	\caption{Comparison of model sizes for different methods.}
	\label{tab:comparison-para}
	\begin{tabular}{lr}
		\toprule
		Model      & \# Parameters      \\
		\midrule
		BicycleGAN &          7.91M     \\
		MS-Pix2Pix &  \textbf{7.61}M    \\
		TET-GAN    &          124.39M   \\
		MC-GAN     &          76.42M    \\
		\textbf{AGIS-Net}    & 19.93M   \\
		\bottomrule
	\end{tabular}

\end{table}



\begin{figure}[t]
	\centering
	\includegraphics[width=0.9\columnwidth]{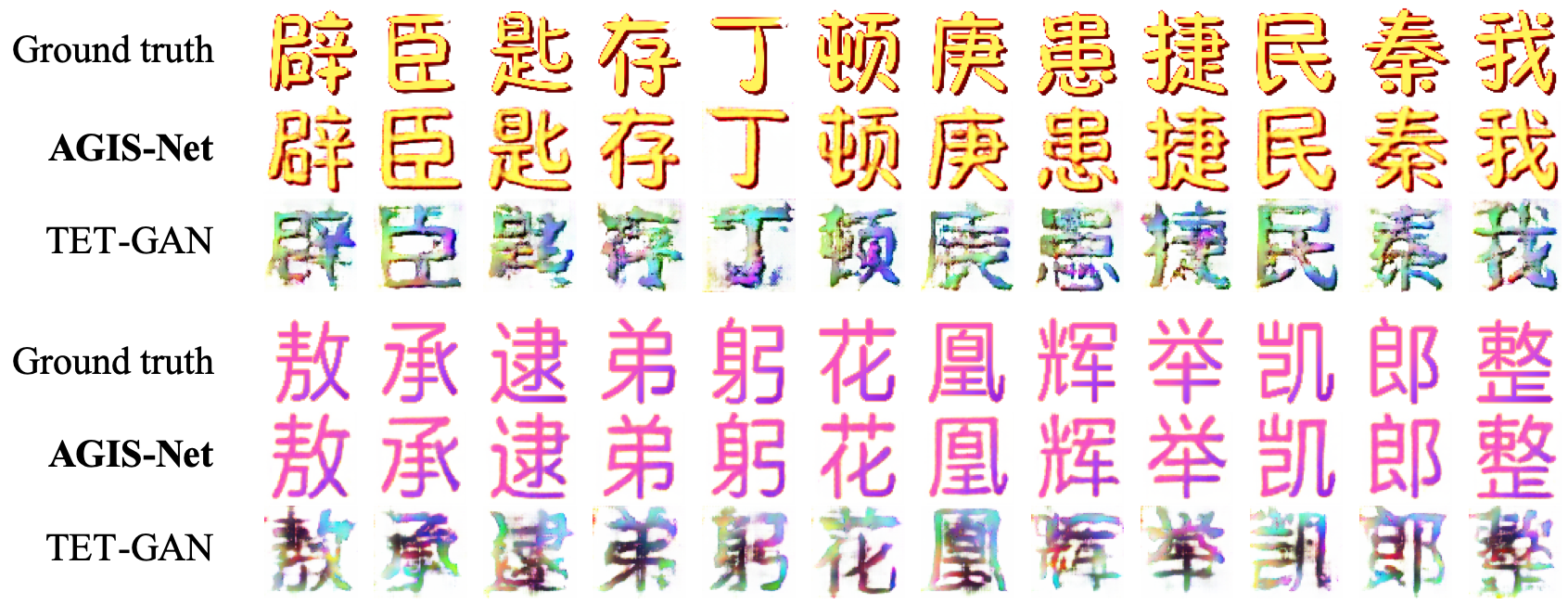}
	\caption{\label{fig:chinese-res}
		Some results of our proposed AGIS-Net and TET-GAN~\cite{yang2019tet} on few-shot learning for Chinese glyph images. All these characters are not in the few-shot reference set.}
\end{figure}

\begin{table}
	\centering
	\caption{Quantitative comparison of our AGIS-Net and TET-GAN~\cite{yang2019tet} on the Chinese glyph image dataset.}
	\label{tab:comparison-cn}
	\begin{tabular}{lccccc}
		\toprule
		Model               & IS              & FID               & SSIM            & pix-acc  \\
		\midrule
		TET-GAN             & \textbf{2.3285} & 165.308           & 0.3937          &  0.4775 \\
		\textbf{AGIS-Net}    & 2.1122          & \textbf{70.875}   & \textbf{0.6116} &  \textbf{0.7035}   \\
		\bottomrule
	\end{tabular}

\end{table}

\subsection{Application on Chinese Glyph Image Synthesis}
As mentioned before, our method is suitable to handle the artistic glyph image synthesis task for any writing systems. Both the style input and content input are flexibly controllable. In this section, we conduct experiments on the Chinese glyph image dataset to verify the extendibility of our AGIS-Net.

In Figure~\ref{fig:chinese-res} and Table \ref{tab:comparison-cn}, we show some experimental results of different methods trained and tested on the Chinese dataset. Due to the fact that MC-GAN~\cite{azadi2018multi} can only handle 26 Latin capital letters, here we just compare our method with TET-GAN~\cite{yang2019tet}. We can see that glyph images synthesized by our method not only precisely inherit the corresponding font's overall and detailed styles but also clearly represent the correct contents of characters. On the contrary, neither the shape style nor texture style can be satisfactorily transferred by TET-GAN, indicating that our method performs significantly better than TET-GAN in this specific task.


\begin{figure}[t]
	\centering
	\includegraphics[width=0.9\columnwidth]{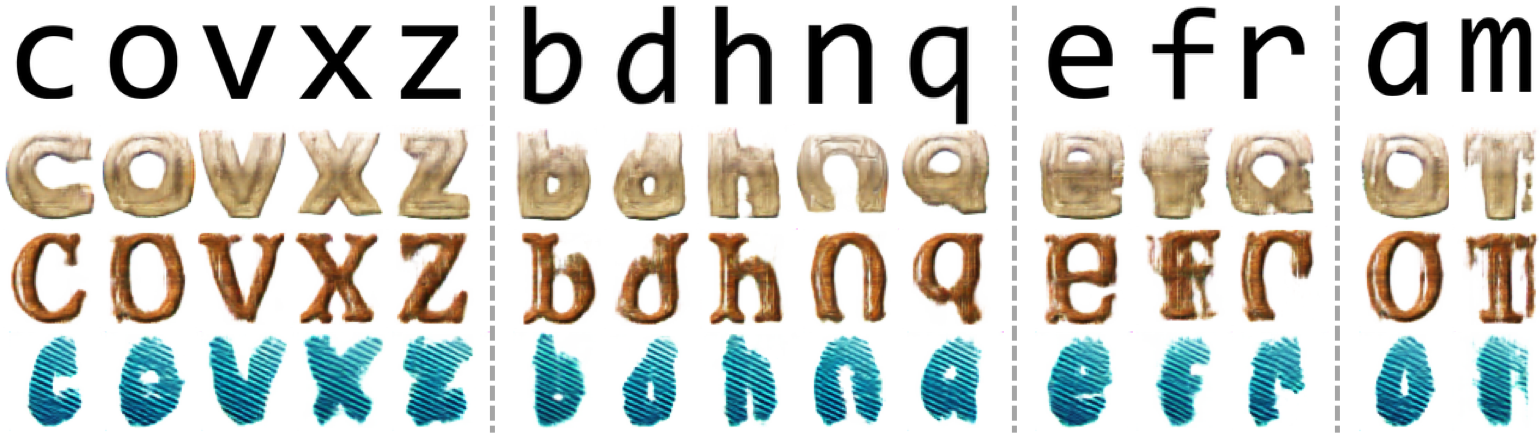}
	\caption{\label{fig:lower}
		Glyph image synthesis for English lowercase characters. The first row shows the content input for lowercase characters.}
\end{figure}

\begin{figure}[t]
	\centering
	\includegraphics[width=0.95\columnwidth]{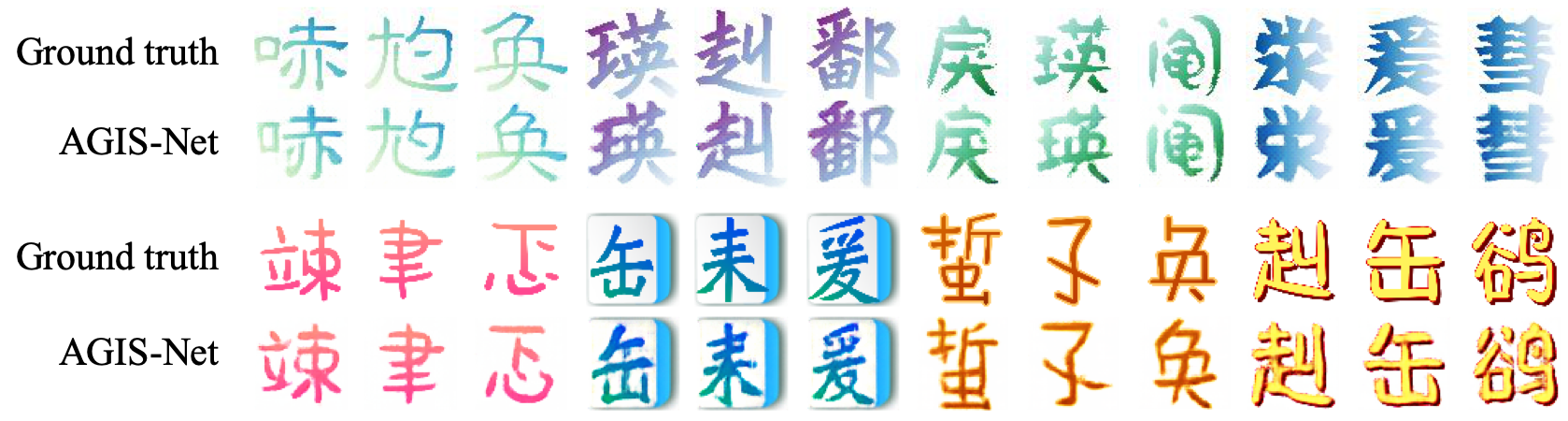}
	\caption{\label{fig:cn-unexplored}
		Glyph image synthesis for unexplored Chinese characters, these characters are unseen during both pre-training and few-shot learning.}
\end{figure}


\subsection{Generalization Ability}
To verify the generalization ability of our model, we conduct the following two experiments.

\begin{figure}[t]
	\centering
	\includegraphics[width=0.95\columnwidth]{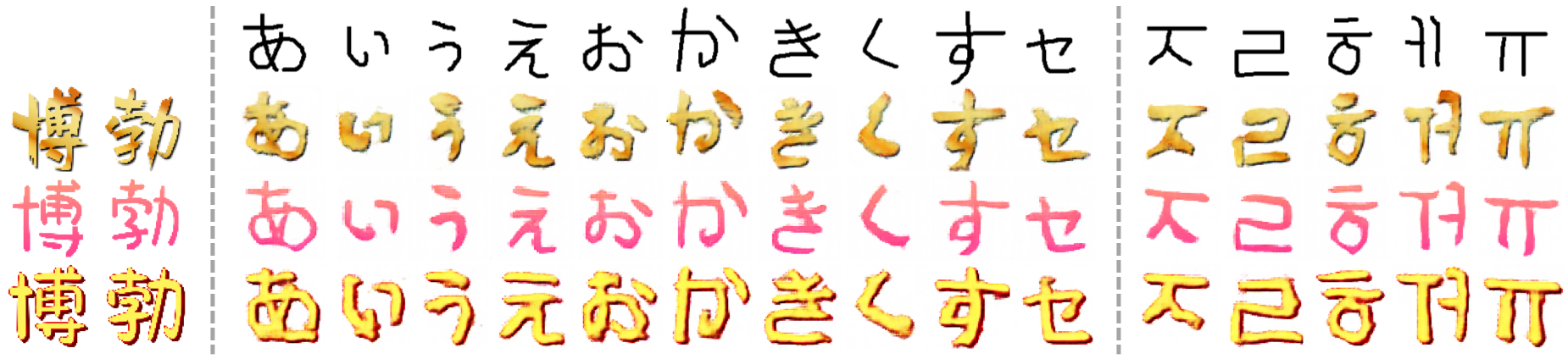}
	\caption{\label{fig:across}
		Glyph image synthesis for Japanese and Korean characters using our learnt models trained on Chinese glyph images (see some samples in the leftmost two columns). The first row shows the content input images.}
\end{figure}

\subsubsection{Synthesizing unexplored glyph images}
We conduct this experiment to show that besides the glyphs explored in training, our model also has the ability of generating high-quality glyph images for unexplored characters, which are unseen during both pre-training and few-shot learning procedures. For the English dataset, all lowercase letters are unexplored. As we can see from Figure \ref{fig:lower}, for the lowercase English characters (the first part) which are similar to their uppercase counterparts, our model can generate satisfactory results. For most lowercase letters (the second part) which are quite different from their uppercase versions, our model can also generate reasonable results. For some characters (the third part), our model only works well for some styles. As shown in the last part of Figure \ref{fig:lower}, our method fails when handling two lowercase letters: 'a' and 'm'. For 'a', which is quite similar to 'o' in this content input font style, our model tends to synthesize results like 'o' for all styles. For 'm', incorrect synthesis results are obtained due to its unique shape structure with 3 vertical lines that is quite different compared to other characters.

A similar experiment is also conducted on the Chinese dataset, as shown in Figure \ref{fig:cn-unexplored}, the styles of glyphs in the first two rows come from the pre-training set, where glyph images of 500 characters in different styles are used to train our model, and the styles of glyphs in the last two rows come from the fine-tuning set, where only 30 stylized glyph images are available for training. Then, we input glyph images of some content reference characters that are unseen in both pre-training and fine-tuning procedures, and get corresponding synthesized glyph images. Surprisingly, the quality of synthesis results for these unexplored characters is comparable to that of explored ones (see Figure~\ref{fig:cn-unexplored}).

\subsubsection{Cross-language evaluation}
The idea of cross-language evaluation is similar to the former experiment. We pre-train our model on the Chinese glyph image dataset and fine-tune the model on specific Chinese artistic font styles. Then, we feed the learnt model with Japanese and Korean glyph images on the Omniglot dataset~\cite{Lake2015HumanlevelCL} as content input. As we can see from Figure \ref{fig:across}, synthesis results of our model for Japanese/Korean characters are also impressive.

Experimental results demonstrate that our model has powerful generalization ability, and is potentially suitable to handle many other relevant tasks (e.g., painting synthesis). These results also verify that the proposed model is capable of disentangling the content and style for glyph images.


\begin{figure}[t]
	\centering
	\includegraphics[width=0.9\columnwidth]{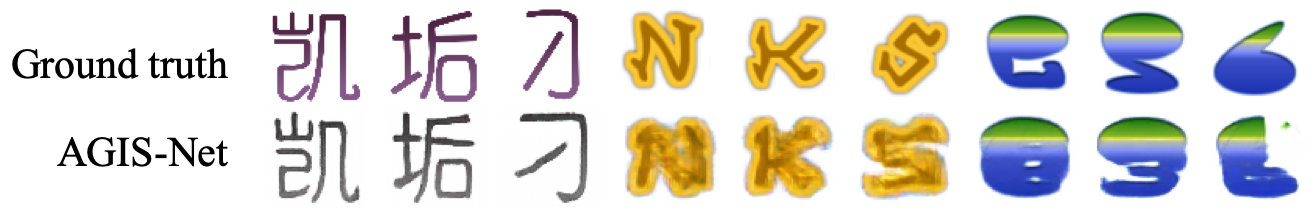}
	\caption{\label{fig:fail}
		Failure cases of our AGIS-Net for some artistic font styles.}
\end{figure}

\subsection{Failure Cases}
Our model sometimes fails in generating satisfactory synthesis results. As shown in Figure~\ref{fig:fail}, the quality of generated Chinese glyph images in the first three columns are poor due to the color inconsistency, probably because the model falls into a local minimum during training, this could be fixed by tuning the hyper-parameters. Our model also does not perform well for the other two English fonts, as their shape styles are so unique that there exists a huge style gap between them and the pre-training data.


\begin{figure}[t]
	\centering
	\includegraphics[width=\columnwidth]{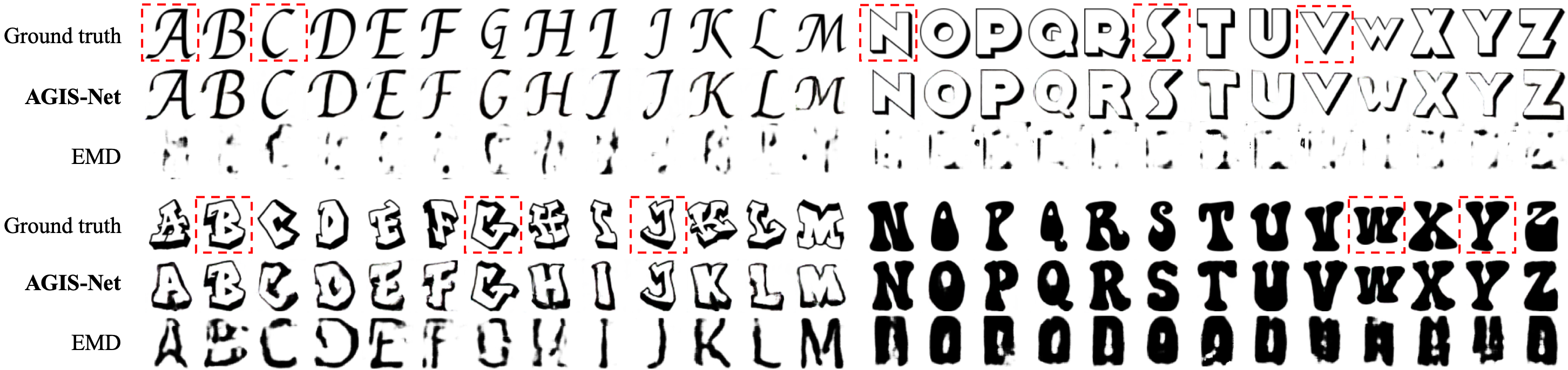}
	\caption{\label{fig:shape_transfer}
		Shape glyph image synthesis results of our model and EMD~\cite{zhang2018separating}, characters in the few-shot reference set are marked in red boxes.}
\end{figure}

\subsection{Analysis on Shape Style Transfer}

The task of shape style transfer is much tougher than texture style transfer. To see how the shape decoder performs, we remove the texture decoder and $D_{tex}$ from our model. Here we compare our model with a recently proposed method EMD~\cite{zhang2018separating} which also aims at separating content and style. We select several fonts from the test set of the English pre-training dataset for the few-shot learning task. As shown in Figure~\ref{fig:shape_transfer}, our model clearly outperforms EMD on all these fonts, indicating the effectiveness and superiority of our model for shape style transfer. However, there also exist some failure cases when using our model for glyph shape style transfer. For example, our synthesis results (e.g., 'A', 'M' and 'Z') shown in the fifth row of Figure~\ref{fig:shape_transfer} look obviously different compared to their corresponding ground truth shapes. Actually, not only the proposed model but also all other existing approaches can not satisfactorily transfer the styles of some fonts, in which most characters have their own unique shape style or/and local details.

\section{Conclusion}

In this paper, we proposed a novel one-stage few-shot learning model for artistic glyph image synthesis. The proposed AGIS-Net only needs a small number of training samples as input, and then high-quality glyph images in the same artistic style as training data can be synthesized for any other characters. We also built a new large-scale Chinese glyph image dataset for performance evaluation. Experiments on two publicly available datasets demonstrate that our model is capable of generating high-quality bitmap images of characters while maintaining content information and style consistency. It should be pointed out that those bitmap images cannot be directly used to create a font consisting of vector images of characters which are perfectly scalable. How to automatically synthesize an artistic font that contains large numbers of glyph vector images using just a few samples is an interesting and challenging direction for future research.


\begin{acks}

This work was supported by National Natural Science Foundation of China (Grant No.: 61672043 and 61672056),  Center For Chinese Font Design and Research, and Key Laboratory of Science, Technology and Standard in Press Industry (Key Laboratory of Intelligent Press Media Technology).

\end{acks}

%
%
%
%


\bibliographystyle{ACM-Reference-Format}
\bibliography{bibliography}

%

\end{document}